\documentclass[a4paper,fleqn]{cas-dc}

\usepackage[numbers,sort&compress]{natbib}
\usepackage{graphicx}
\usepackage{booktabs}
\usepackage{amsmath}
\usepackage{amssymb}
\usepackage{multirow}
\usepackage{microtype}

\usepackage{cleveref}
\usepackage{algorithm}
\usepackage{algorithmic}

\newcommand{\R}{\mathbb{R}}
\newcommand{\E}{\mathbb{E}}
\newcommand{\N}{\mathcal{N}}

\newcommand{\X}{\mathbf{X}}                 
\newcommand{\hatX}{\hat{\mathbf{X}}}        
\newcommand{\Yvec}{\mathbf{Y}}              

\newcommand{\Usys}{U_{\mathrm{sys}}}        
\newcommand{\Udepol}{U_{\mathrm{depol}}}
\newcommand{\Urepol}{U_{\mathrm{repol}}}
\newcommand{\Uenv}{U_{\mathrm{env}}}

\newcommand{\Zone}{Z_1}                     
\newcommand{\Zperf}{Z_{\mathrm{perf}}}      
\newcommand{\Zstruc}{Z_{\mathrm{struc}}}    
\newcommand{\Zcond}{Z_{\mathrm{cond}}}      
\newcommand{\Zrhythm}{Z_{\mathrm{rhythm}}}  

\newcommand{\Zdepol}{Z_{2,\mathrm{depol}}}
\newcommand{\Zrepol}{Z_{2,\mathrm{repol}}}

\newcommand{\Zqrs}{Z_{3,\mathrm{qrs}}}
\newcommand{\Zstt}{Z_{3,\mathrm{stt}}}

\newcommand{\Lrecon}{\mathcal{L}_{\mathrm{recon}}}
\newcommand{\Lkl}{\mathcal{L}_{\mathrm{KL}}}
\newcommand{\Ltask}{\mathcal{L}_{\mathrm{task}}}
\newcommand{\Ltopo}{\mathcal{L}_{\mathrm{topo}}}
\newcommand{\Rtopo}{\mathcal{R}_{\mathrm{topo}}}
\newcommand{\Ltotal}{\mathcal{L}_{\mathrm{total}}}

\newcommand{\lambdatask}{\lambda_{\mathrm{task}}}
\newcommand{\lambdaanchor}{\lambda_{\mathrm{anchor}}}
\newcommand{\lambdaKL}{\beta}

\newcommand{\gdec}{g_{\mathrm{dec}}}        
\newcommand{\fdepol}{f_{\mathrm{depol}}}
\newcommand{\frepol}{f_{\mathrm{repol}}}
\newcommand{\fqrs}{f_{\mathrm{qrs}}}
\newcommand{\fstt}{f_{\mathrm{stt}}}

\begin{document}

\let\WriteBookmarks\relax
\def\floatpagepagefraction{1}
\def\textpagefraction{.001}

\shorttitle{TRACE: Tractable Routing Autoencoder for Clinical ECG}
\shortauthors{}

\title [mode = title]{TRACE: Tractable Routing Autoencoder for Clinical ECG}

\author[1]{Shunbo Jia}[orcid=0009-0003-9854-9930]
\fnmark[1]
\ead{2240003657@student.must.edu.mo}
\credit{Conceptualization, Methodology, Investigation, Formal analysis}

\affiliation[1]{organization={Department of Bioelectronics, Faculty of Biomedical Engineering, Shenzhen University of Advanced Technology},
            city={Shenzhen},
            postcode={518107},
            country={China}}

\author[2]{Runze Ma}[orcid=0009-0001-2620-0575]
\fnmark[1]
\ead{rmaa0033@student.monash.edu}
\credit{Software, Investigation, Visualization}

\affiliation[2]{organization={Faculty of Information Technology, Monash University},
            city={Clayton},
            state={Victoria},
            postcode={3800},
            country={Australia}}

\author[1]{Haonan Lyu}[orcid=0009-0005-2932-816X]
\ead{SUAT25060153@stu.suat-sz.edu.cn}
\credit{Investigation, Data curation}

\author[3]{Haijin Zhang}[orcid=0009-0009-9124-2446]
\ead{haijin.zhang@student.uq.edu.au}
\credit{Validation, Formal analysis}

\affiliation[3]{organization={Faculty of Science, University of Queensland},
            city={Brisbane},
            postcode={4072},
            country={Australia}}

\author[4]{Qiang Yang}[orcid=0000-0002-5202-7892]
\cormark[1]
\ead{qy258@cam.ac.uk}
\credit{Supervision}

\affiliation[4]{organization={Department of Computer Science and Technology, University of Cambridge},
            city={Cambridge},
            postcode={CB2 1TN},
            country={UK}}

\author[1]{Caizhi Liao}[orcid=0000-0003-1271-3027]
\cormark[1]
\ead{liaocaizhi@suat-sz.edu.cn}
\credit{Supervision, Project administration}

\cortext[1]{Corresponding author}
\fntext[1]{These authors contributed equally to this work.}

\begin{abstract}
Deep learning has advanced automated electrocardiogram (ECG) diagnosis, but the field's most accurate models, foundation models pretrained on millions of recordings, are not decision-pathway auditable: a clinician cannot trace a diagnosis to a physiological pathway or intervene on one. We propose TRACE, a \textbf{T}ractable \textbf{R}outing \textbf{A}utoencoder for \textbf{C}linical \textbf{E}CG, whose 32-dimensional clinical latent space is specified in advance from domain knowledge rather than discovered by optimization. TRACE partitions this space into perfusion, structure, and conduction subspaces, routes each to its own diagnostic head by design, regularizes the partition with an orthogonality penalty, and reconstructs the ECG through a decoder that permits latent perturbation. On PTB-XL and Georgia, TRACE exceeds unconstrained classifiers and stays ahead of an ECG foundation model pretrained on ten million recordings, evaluated by linear probe on frozen features, at roughly an eighth of the parameter count. On the nine-label CPSC2018 cohort, which carries no structural class, the framework transfers with only the routing table re-specified to a perfusion/rhythm/conduction partition. Joint probe, erasure, and perturbation analyses verify the routing contract, and perturbing the depolarization and repolarization pathways modulates the reconstructed waveform. Removing the specified partition and its orthogonality penalty costs 1.70 AUC and 11.30 macro-F1 points on PTB-XL, and 2.76 AUC and 16.92 macro-F1 points on Georgia. A capacity-matched permutation control places arbitrary assignments within 0.34 AUC points of the ontology routing and leaves macro-F1 statistically level ($p=0.619$): the ontology supplies decision-pathway auditability at no macro-F1 cost.
\end{abstract}

\newif\ifshowhighlights
\showhighlightsfalse
\ifshowhighlights
\begin{highlights}
\item Routing is fixed by the ontology and re-specifiable at no macro-F1 cost.
\item Outperforms ECG foundation models on PTB-XL and Georgia with far fewer parameters.
\item Probe, erasure, and perturbation analyses verify the routing contract.
\item Removing the partition costs 11.30 and 16.92 macro-F1 points.
\end{highlights}
\fi


\maketitle


\section{Introduction}\label{sec:introduction}

Deep learning has advanced automated electrocardiogram (ECG) diagnosis \cite{siontis2021artificial,hannun2019cardiologist}. These gains, from supervised classifiers to foundation models pretrained on large ECG corpora \cite{ECGFounder2025,HeartLang2025}, have come at the cost of transparency. In both regimes the latent representation is not organized by clinical factor, and a clinician cannot trace a diagnosis to any physiological basis or intervene on a specific pathway. That opacity has clinical consequences. Standard models can rely on statistical shortcuts that fail under distribution shift \cite{d2022underspecification,geirhos2020shortcut}, and clinicians are rightly reluctant to trust predictions whose internal reasoning they cannot inspect \cite{rudin2019stop,ghassemi2021false}.

Concept bottleneck models \cite{koh2020concept} make the intermediate quantity readable where annotated concepts exist, but that readability sits at the label level, no clinical pathway is separated, and accuracy often drops relative to unconstrained baselines. Post-hoc explanations report what a fitted model did without constraining what it can do \cite{rudin2019stop}. Both routes leave cardiology's knowledge of how the ECG is generated outside the model's structure.

This study follows a different route: where reliable domain knowledge exists but the supervision needed to learn it does not, we build that knowledge into the model's skeleton. We propose TRACE, a \textbf{T}ractable \textbf{R}outing \textbf{A}utoencoder for \textbf{C}linical \textbf{E}CG diagnosis. Its latent space is \emph{specified in advance from domain knowledge} rather than discovered by optimization. TRACE partitions that space into three named subspaces for perfusion, cardiac structure, and electrical conduction, routes each subspace by design to its own diagnostic head, and regularizes the partition with an orthogonality penalty. The model is accurate, the routing behind its decisions is auditable by construction, and its representations can be examined through reconstruction.

The paper's contributions are:

\begin{itemize}
    \item \textbf{A partitioned ECG autoencoder with fixed routing.} TRACE instantiates a literature-derived diagnostic ontology (Section~\ref{sec:preliminaries}) as a concrete routing contract: partitioned latent space, isolated heads, and an orthogonality penalty, with the ontology as the part that can be re-specified. On PTB-XL and Georgia it exceeds unconstrained classifiers and open-weight ECG foundation models, from scratch, on orders-of-magnitude fewer recordings and at a fraction of their parameter count (Section~\ref{sec:main-results}).
    \item \textbf{A routing-contract verification protocol.} We evaluate structured-representation models by reading probe, erasure, and perturbation analyses jointly, and we release the suite with our code. Erasure mechanically confirms the wiring, and probing measures the cross-subspace information the construction leaves open (Section~\ref{sec:routing-verification}).
    \item \textbf{Representation-level evidence.} Beyond routing, the reconstruction pathway preserves waveform morphology (Pearson $r = 0.905$), and interventions on the depolarization and repolarization pathways (the mechanism layer) modulate the reconstructed waveform. A three-seed, full-test-set protocol characterizes that modulation at the granularity of that layer (Section~\ref{sec:representation}).
    \item \textbf{Robustness under realistic perturbations.} The knowledge-specified design resists electrode noise and lead masking without adversarial training or perturbation-specific augmentation (Section~\ref{sec:robustness}).
\end{itemize}

The rest of the paper is organized as follows. Section~\ref{sec:related} positions TRACE against deep ECG classification, ECG foundation models, concept-based interpretability, and knowledge-guided representation learning. The clinical knowledge base that TRACE instantiates is set out in Section~\ref{sec:preliminaries}, and Section~\ref{sec:method} turns it into an architecture. Section~\ref{sec:experiments} covers diagnostic performance and the routing-contract verification, then ablations, representation-level evidence, and robustness. Section~\ref{sec:discussion} examines the value of routing-level transparency, the boundary between specified and learned structure, and what observational evidence can establish in ECG research. Section~\ref{sec:conclusion} concludes.


\section{Related Work}\label{sec:related}

\subsection{Deep ECG diagnosis: from supervised classifiers to foundation models}

The accuracy line of work in ECG deep learning began with adapted convolutional and attention architectures \cite{hannun2019cardiologist,ribeiro2020automatic,wang2017time,ismail2020inceptiontime,vaswani2017attention}, including designs that encode lead structure explicitly rather than leaving it to be discovered \cite{li2023dualscale}. It then moved to self-supervised pretraining \cite{kiyasseh2021clocs,chen2020simple,mehari2022self,he2022masked}, with masked-autoencoder families adapted to ECG \cite{zhang2023maefe,pham2025maskedecgtext} and pretraining on large unlabeled biosignal corpora \cite{abbaspourazad2024wearable}. Most recently it has moved to foundation models trained on very large corpora: ECGFounder, with 150-class expert-annotated pretraining on over ten million recordings \cite{ECGFounder2025}, and HeartLang, with language-inspired tokenization \cite{HeartLang2025}. Open-weight alternatives \cite{mckeen2025ecgfm}, vision-transformer variants \cite{vaid2023foundational}, physiology-informed designs whose objective rests on a differentiable hemodynamic model \cite{papastathopoulos2026physics}, and spatio-temporal masking designs such as ST-MEM \cite{na2024guiding} extend the same scaling strategy. Each step improves benchmark accuracy while leaving the same problem intact: a linear probe can report how accurately a diagnosis is decoded from the representation, but not which physiological pathway produced it. TRACE shares the supervised backbone of this line of work and replaces the unconstrained latent space with an ontology-specified one. The comparison in Section~\ref{sec:main-results} therefore measures what the specification adds in accuracy.

\subsection{Interpretability and concept-based models}

The transparency line of work addresses opacity from outside the model. Concept bottleneck models force predictions through human-defined concepts. They require concept annotations, can trade accuracy for transparency \cite{koh2020concept}, and the concepts they learn need not be the ones intended \cite{margeloiu2021concept}. Variants relax the annotation requirement by learning concepts without labels \cite{oikarinen2023labelfree}, by embedding concepts in higher-dimensional spaces \cite{espinosazarlenga2022concept}, or through energy-based formulations that support intervention \cite{xu2024energy}. Prototype-based models instead make the evidence for a prediction inspectable by construction, matching learned prototypical parts \cite{nauta2023pipnet}, an approach recently adapted to multi-label ECG classification \cite{sethi2025protoecgnet}. Probing methods supply evidence about what a representation contains, from observational correlation to causal erasure \cite{elazar2021amnesic,belinkov2022probing}. Attention-based ECG architectures provide per-recording attention maps \cite{yisimitila2025interpretable}, and PTB-XL benchmark studies document interpretability properties of ECG models \cite{strodthoff2021deep}. A recent systematic review surveys the wider explainable-AI literature for electrocardiography \cite{taleban2026explainable}. Two gaps recur. First, post-hoc explanations describe model behavior without constraining it, so the explanation need not correspond to the decision path actually taken \cite{rudin2019stop}. Second, concept supervision locates interpretability at the label level, outside the model's internal routing between physiological subsystems. TRACE addresses the first gap by making routing a hard architectural constraint and the second by verifying the routing contract directly (Section~\ref{sec:routing-verification}).

\subsection{Knowledge-guided and structured representation learning}

Knowledge-guided work embeds domain structure into the representation itself. Grounding predictions in domain knowledge and structured modeling assumptions improves robustness across medical and scientific settings \cite{subbaswamy2020development}. Surveys of knowledge-guided machine learning map how scientific knowledge is typed, integrated and incorporated \cite{karpatne2024knowledgeguided}. Logic-based frameworks integrate first-order rules into a trainable network \cite{badreddine2022logic}, and physics-informed networks carry governing equations into the training objective itself \cite{raissi2019physics}. Disentanglement methods seek independent generative factors through restricted-capacity bottlenecks or total-correlation penalties \cite{higgins2017betavae,kim2018disentangling,chen2018isolating}, with analyses of learned factors under correlated data \cite{trauble2021disentangled} and causal perspectives on representation learning \cite{scholkopf2021toward}. This line of work also bounds what unsupervised discovery can establish. Disentangled representations are unidentifiable without inductive bias on both models and data \cite{locatello2019challenging}, and identification becomes possible only when the prior is conditioned on additional observed structure \cite{khemakhem2020variational}. A parallel statistical tradition models biomedical signals as functional data, with functional regression for physiological prediction and multilevel distributional models for wearable biosensor streams \cite{matabuena2019heartrate,matabuena2024multilevel,ghosal2025functional}. TRACE's partitioned latent space and its patient-level context are neural counterparts of that tradition's structured assumptions, and its reconstruction pathway is the fidelity check. TRACE joins this line of work by specifying the factorization rather than discovering it. The partition, its dimensions, and its routing are fixed from the diagnostic ontology of Section~\ref{sec:ontology} before training, and learning fills in the content of each subspace under those constraints. As Section~\ref{sec:ablations} shows, the specified structure carries the model's multi-label performance. To our knowledge, no prior ECG work combines an ontology-specified clinical latent partition with hard routing, an orthogonality penalty, and a reconstruction pathway verified by probing, erasure, and perturbation analyses read jointly.


\section{Preliminaries: The Clinical Knowledge Base}\label{sec:preliminaries}

A model whose structure rests on domain knowledge owes the reader an explicit statement of that knowledge. This section sets out the two knowledge sources that TRACE instantiates as architecture: the electrophysiology of the ECG waveform and a diagnostic ontology that maps clinical categories to waveform evidence. The section closes with the notation used throughout the paper.

\subsection{Cardiac electrophysiology and the 12-lead ECG}
\label{sec:electrophysiology}

The 12-lead ECG records the body-surface projection of cardiac electrical activity \cite{marriott2008practical,guyton2020textbook}. Each cardiac cycle comprises two electrophysiologically distinct processes. \emph{Depolarization}, the sequential activation of atrial and ventricular myocardium, produces the P wave and the QRS complex \cite{guyton2020textbook}. \emph{Repolarization}, the recovery of the myocardium to its resting state, produces the ST segment and the T wave \cite{marriott2008practical,wellens200640}. Different ion-channel systems govern the two processes, and pathology affects them differently \cite{guyton2020textbook}.

This dichotomy organizes clinical ECG interpretation. Structural and conduction abnormalities manifest predominantly in depolarization-derived components: increased muscle mass raises QRS voltage \cite{braunwald2018heart}, and delayed or interrupted propagation widens the QRS complex and alters its activation sequence \cite{josephson2015clinical}. Ischemia and infarction manifest predominantly in repolarization-derived components: inadequate coronary flow shifts the ST segment and inverts the T wave \cite{thygesen2012universal,wellens200640}. TRACE uses this waveform-level division as the backbone of its latent structure. Encoding such structure explicitly has precedent in ECG architectures that operate on the leads separately \cite{li2023dualscale} and in physiological models whose constraints enter the training objective through a differentiable simulator \cite{papastathopoulos2026physics}. Neither specifies the diagnostic partition itself, which is the step TRACE takes here.

\subsection{The diagnostic ontology used by TRACE}
\label{sec:ontology}

Table~\ref{tab:ontology} states the diagnostic ontology that TRACE instantiates. The table maps each diagnostic category to its conventional waveform evidence, the electrophysiological process that produces it, and a designated latent subspace. Three remarks clarify the status of this ontology.

\begin{table*}[!ht]
\caption{The diagnostic ontology used by TRACE, mapping each diagnostic category to its waveform evidence, electrophysiological process, and designated latent subspace.}
\label{tab:ontology}
\begin{tabular*}{\tblwidth}{@{}LLLL@{}}
\toprule
Diagnostic category & Waveform evidence & Process & Subspace \\
\midrule
ST--T changes (STTC) & ST-segment deviation, T-wave inversion & Repolarization & $\Zperf$ \\
Myocardial infarction (MI) & pathological Q waves, ST deviation & Depol.\ and repol. & $\Zperf$ \\
Hypertrophy (HYP) & increased QRS voltage & Depolarization & $\Zstruc$ \\
Conduction disturbance (CD) & widened QRS, altered activation & Depolarization & $\Zcond$ \\
Normal (NORM) & no diagnostic waveform change & --- & all of $\Zone$ \\
\bottomrule
\end{tabular*}
\end{table*}

First, the ontology is a \emph{clinical convention} distilled from textbook electrocardiography \cite{marriott2008practical,braunwald2018heart,josephson2015clinical,thygesen2012universal}. TRACE uses it as \emph{specified knowledge}: it fixes the latent partition and the routing of prediction heads before training. Section~\ref{sec:method} details how the ontology becomes architecture, and Section~\ref{sec:discussion} returns to the boundary between specified structure and discovered structure.

Second, the ontology is adapted per dataset to match label availability while preserving the underlying waveform logic. PTB-XL \cite{wagner2020ptbxl} uses the five-category ontology of Table~\ref{tab:ontology} directly, and Georgia \cite{perezalday2020challenge,perezalday2022challengephysionet} is mapped onto it from its native SNOMED-CT annotations. CPSC2018 \cite{liu2018open} carries a nine-label set with no structural class. There the structure subspace is replaced by a rhythm subspace $\Zrhythm$~serving atrial-fibrillation and premature-complex labels, while the perfusion and conduction subspaces retain their semantics. Table~\ref{tab:routing_app} summarizes the resulting routing per dataset. $\Zrhythm$ captures the rhythm disorders AF, PAC, and PVC, which alter the activation sequence and are distinct from conduction blocks. On every dataset the NORM head receives the full concatenated $\Zone$, because normality is the absence of abnormality across all subsystems.

Third, the number of subspaces follows the clinically meaningful evidence axes that the label system carries. Two categories share a subspace when they share both waveform evidence and electrophysiological process: MI and STTC both carry ST-segment evidence. They split when their evidence differs under a shared process: HYP raises QRS voltage while CD widens the QRS, both depolarization-derived but with distinct evidence. An evidence axis with no supervised labels would collapse to noise, which is why $\Zstruc$ disappears on CPSC2018 and a rhythm axis takes its place. The design permits more axes; Section~\ref{sec:ablations} quantifies what an additional axis would buy. The three-axis grouping is the finest clinically meaningful granularity the label system supports. The number of axes is a property of the label system, and the naming of each axis is a normative clinical specification. The dimensionality of each axis is a hyperparameter whose minimum the sensitivity sweep of Section~\ref{sec:ablations} places at eight.

\emph{Routing-level transparency} is an architectural guarantee: because each prediction head reads only its designated subspace, the routing behind every diagnosis is auditable by construction. \emph{Representation-level evidence} is empirical, and Section~\ref{sec:routing-verification} obtains it from probing, erasure, and perturbation analyses.

\begin{table}[!ht]
\caption{Semantic routing configuration across the three datasets, listing each latent subspace and the labels routed to it.}
\label{tab:routing_app}
\begin{tabular*}{\tblwidth}{@{}LLL@{}}
\toprule
Dataset & Latent subspace & Routed labels \\
\midrule
PTB-XL / Georgia & $\Zperf \in \R^{16}$ & MI, STTC \\
PTB-XL / Georgia & $\Zstruc \in \R^{8}$ & HYP \\
PTB-XL / Georgia & $\Zcond \in \R^{8}$ & CD \\
\cmidrule(lr){1-3}
CPSC2018 & $\Zperf \in \R^{16}$ & STD, STE \\
CPSC2018 & $\Zrhythm \in \R^{8}$ & AF, PAC, PVC \\
CPSC2018 & $\Zcond \in \R^{8}$ & I-AVB, LBBB, RBBB \\
\bottomrule
\end{tabular*}
\end{table}

\subsection{Problem formulation and notation}
\label{sec:notation}

We study multi-label ECG classification, where each recording is a fixed-length segment $\X \in \R^{C \times T}$ with $C = 12$ leads and $T = 5000$ samples at $500$\,Hz. The preprocessing pipeline is shared by all models in this study (Section~\ref{sec:setup}). The label vector is $\Yvec \in \{0,1\}^{K}$ over $K$ diagnostic categories; multiple categories can be positive for the same recording, since clinical co-morbidity is common.

TRACE maps $\X$ to predictions through a hierarchical latent representation summarized in Table~\ref{tab:notation}. The central structural choice is the clinical latent space $\Zone = [\Zperf; \Zstruc; \Zcond]$, partitioned into three subspaces of 16, 8, and 8 dimensions; throughout, $\psi_k$ denotes the prediction head of category $k$ and $r(k)$ the subspace the ontology assigns to it. Section~\ref{sec:method} describes the architecture in full.

\begin{table*}[!ht]
\caption{Notation used throughout the paper, with the dimensionality of each quantity where applicable.}
\label{tab:notation}
\begin{tabular*}{\tblwidth}{@{}LLL@{}}
\toprule
Symbol & Meaning & Dimensionality \\
\midrule
$\X$ & input 12-lead ECG segment & $12 \times 5000$ \\
$\Yvec$ & multi-label diagnostic vector & $K$ \\
$h$ & global temporal feature & $256 \times 157$ \\
$\Phi_\theta$ & perception (residual) backbone & --- \\
$\Usys$ & global patient-level representation & 128 \\
$\mu_{\theta}$, $\sigma_{\theta}^{2}$ & amortized inference functions (system inference network) & --- \\
$\Zone$ & clinical latent space & 32 \\
$\Zperf$ & perfusion subspace & 16 \\
$\Zstruc$ & structure subspace (PTB-XL, Georgia) & 8 \\
$\Zcond$ & conduction subspace & 8 \\
$\Zrhythm$ & rhythm subspace (CPSC2018) & 8 \\
$\Zdepol$, $\Zrepol$ & depolarization / repolarization pathways & --- \\
$\Udepol$, $\Urepol$, $\Uenv$ & individual-specific and environment additive terms & 64 \\
$\Zqrs$, $\Zstt$ & QRS / ST--T waveform latents & --- \\
$\hatX$ & reconstructed 12-lead ECG & $12 \times 5000$ \\
$f_{*}$, $\gdec$ & learned intermediate mappings and decoder & --- \\
$\psi_k$, $r(k)$ & prediction head and its routed subspace & --- \\
$q_\phi$, $p_\theta$ & amortized posterior and generative model & --- \\
$\Lrecon$, $\Lkl$, $\Ltask$ & reconstruction, KL, and task loss terms & --- \\
$\Rtopo$ ($\Ltopo$) & orthogonality (topology) penalty and its loss term & --- \\
$\lambdaKL$, $\lambdatask$, $\lambdaanchor$ & KL, task, and anchor objective weights & --- \\
\bottomrule
\end{tabular*}
\end{table*}


\section{Method: A Knowledge-Specified Structured Autoencoder}\label{sec:method}

\begin{figure*}[!ht]
    \centering
    \includegraphics[width=\textwidth]{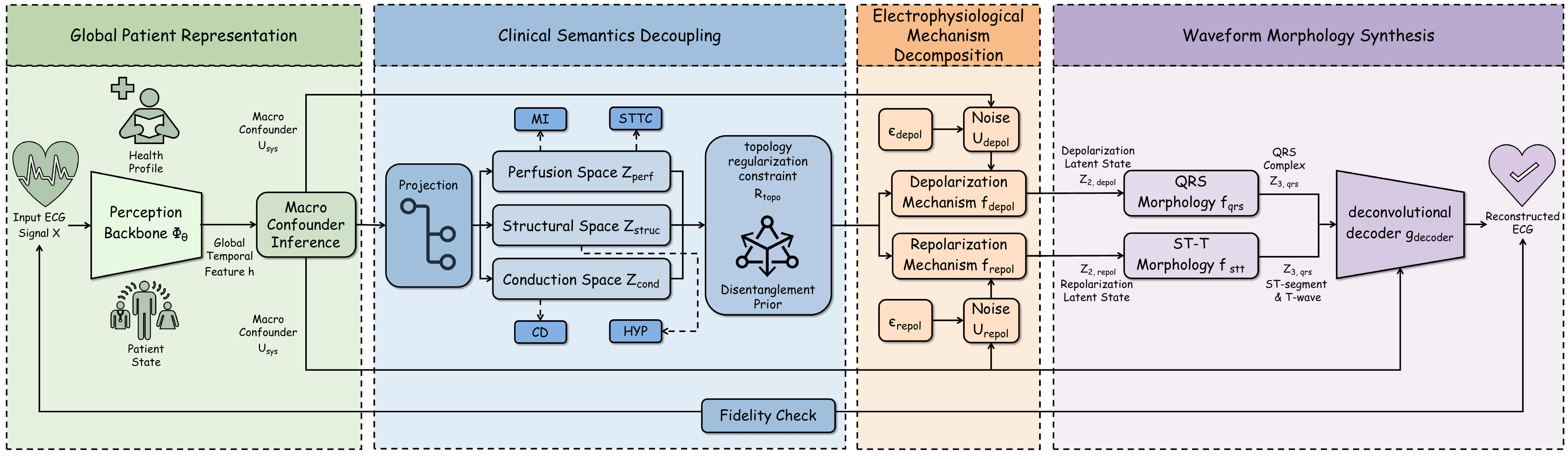}
    \caption{TRACE architecture. Left, perception: a global temporal feature $h$ and the inferred patient-level representation $\Usys$. Center, separation of the clinical subspaces: the partitioned clinical latent space $\Zone=[\Zperf;\Zstruc;\Zcond]$ with isolated diagnostic heads for the perfusion subspace MI and STTC, the structure subspace HYP, and the conduction subspace CD, plus the orthogonality penalty $\Rtopo$. Right, electrophysiological mechanism decomposition and fidelity: depolarization and repolarization pathways, QRS and ST-T waveform synthesis, and the deconvolutional decoder that reconstructs the ECG as a fidelity check.}
    \label{fig:trace_workflow}
\end{figure*}

TRACE turns the clinical knowledge base of Section~\ref{sec:preliminaries} into architecture;
Figure~\ref{fig:trace_workflow} follows the path from the waveform to the reconstructed signal. The construction rests on three design choices, all made \emph{before} training and all auditable afterwards: (K1) the clinical latent space is partitioned into three named subspaces by design; (K2) each prediction head reads only its designated subspace, a hard routing fixed by the ontology; (K3) an orthogonality penalty decorrelates the subspaces as a statistical prior. Section~\ref{sec:routing-verification} reads these commitments against probe, erasure, and perturbation analyses.

\subsection{Perception backbone and patient-level representation}
\label{sec:backbone}

\begin{figure*}[!ht]
    \centering
    \includegraphics[width=\textwidth]{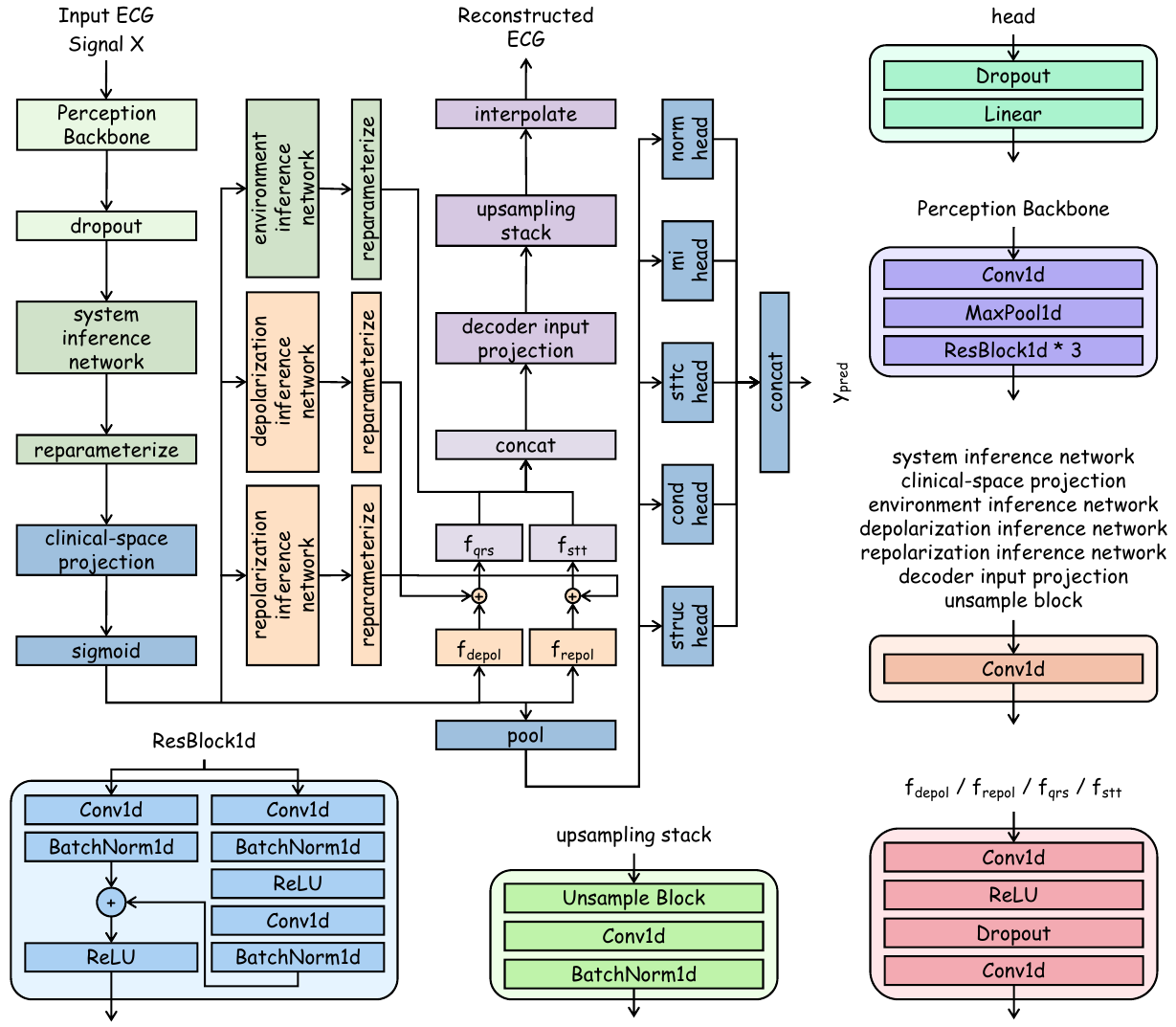}
    \caption{Implementation stack of TRACE, in three regions. Left, the input pathway: the perception backbone, the system inference network with reparameterization, and the clinical-space projection that produces the partitioned latent space $\Zone$. Center, the decomposition pathway: the environment, depolarization, and repolarization inference networks, the $f_{\mathrm{depol}}$/$f_{\mathrm{repol}}$ mappings, the $f_{\mathrm{qrs}}$/$f_{\mathrm{stt}}$ synthesis, and the upsampling decoder. The pooled $\Zone$ feeds the five isolated heads NORM, MI, STTC, CD, and HYP. Detail panels: ResBlock1d, the perception backbone, one head, the upsampling stack, and the $f_*$ mappings.}
    \label{fig:module_stack}
\end{figure*}

A one-dimensional residual backbone $\Phi_\theta$ maps the standardized segment $\X \in \R^{12 \times 5000}$ to a global feature map $h \in \R^{256 \times 157}$; Figure~\ref{fig:module_stack} lays out the concrete layer stack behind every stage described below. From $h$, a global patient-level representation is drawn by amortized variational inference,
\begin{equation}
    \Usys = \mu_{\theta}(h) + \exp\!\left(\tfrac{1}{2}\log \sigma_{\theta}^{2}(h)\right) \odot \epsilon, \qquad \epsilon \sim \N(0,\mathbf{I}),
\end{equation}
with $\Usys \in \R^{128}$. In the implemented model this step is a single \emph{system inference network}: a 15-tap convolution over $h$ whose 256 output channels split into the mean $\mu_{\theta}(h)$ and log-variance $\log\sigma_{\theta}^{2}(h)$ halves, followed by the reparameterization trick \cite{kingma2014autoencoding}. This stochastic bottleneck models patient-level context such as age, chronic conditions, and metabolic state, which modulates how pathology manifests in the waveform. A KL term regularizes $\Usys$ toward a standard Gaussian prior, discouraging the encoder from memorizing patient identity rather than physiological state (Section~\ref{sec:theory} gives the variational derivation).

\subsection{Knowledge-specified clinical latent partition}
\label{sec:partition}

The clinical latent space $\Zone$ is constructed as
\begin{equation}
    \Zone = [\Zperf; \Zstruc; \Zcond] = \sigma\!\left(W_{z_1}\Usys + b_{z_1}\right),
\end{equation}
a 32-dimensional vector partitioned into three subspaces of 16, 8, and 8 dimensions. The mapping is a \emph{clinical-space projection}: a $1\times1$ convolution over $\Usys$ that mixes channels at each time step with no temporal receptive field. Three properties of this partition matter for what follows.

\emph{The partition is specified, not discovered.} The number of subspaces, their dimensions, and their clinical names come from the ontology of Table~\ref{tab:ontology} and from label complexity. The perfusion subspace serves the two diagnostic categories STTC and MI and receives 16 dimensions; the structure and conduction subspaces each serve one category group with 8 dimensions. No component of the training objective learns the assignment, and the specification carries the routing contract of the next paragraph into the architecture. Section~\ref{sec:ablations} varies the dimensions.

\emph{Routing is hard and fixed by the ontology.} Each prediction head $\psi_k$ for category $k$ receives input only from its designated subspace $Z_{1,r(k)}$, with $r$ the routing function of Table~\ref{tab:ontology}. No gradient path connects a head to a non-designated subspace. Whatever the shared encoder learns, a prediction for category $k$ therefore cannot draw on any subspace other than $Z_{1,r(k)}$, and routing-level auditability holds by construction. This hard routing is the primary isolation mechanism, with the penalty below as a complement.

\emph{The partition adapts to the dataset's ontology.} The three axes are fixed, but the clinical content of each follows the corpus's label system: on CPSC2018 the structure subspace gives way to a rhythm subspace $\Zrhythm$ (Section~\ref{sec:ontology}), and Table~\ref{tab:routing_app} lists the resulting routing for all three datasets.

TRACE needs no concept annotations: its subspaces are unsupervised bottlenecks whose routing the ontology fixes.

\subsection{Orthogonality penalty and its statistical interpretation}
\label{sec:orthogonality}

Because perfusion, structural, and conduction abnormalities co-occur through shared pathology \cite{marriott2008practical}, the three subspaces would develop strong correlations if left unregularized. TRACE penalizes pairwise cross-subspace correlation with
\begin{multline}
    \Rtopo = \sum_{i \neq j} \langle Z_{1,i}, Z_{1,j} \rangle^2, \\ Z_{1,i} \in \{\Zperf, \Zstruc, \Zcond\}.
\end{multline}

The penalty admits a statistical reading: minimizing it reduces an approximate upper bound on the pairwise mutual information between subspaces. Section~\ref{sec:theory} states that bound and derives it.

The penalty implements \emph{mini-batch statistical decorrelation}, the statistical prior formalized in Section~\ref{sec:theory}, and reduces redundant cross-subspace leakage in the shared encoder, which the probe matrix of Section~\ref{sec:routing-verification} measures directly.

\subsection{Electrophysiological decomposition and reconstruction}
\label{sec:mechanism}

The clinical subspaces feed two parallel electrophysiological pathways, mirroring the depolarization-repolarization dichotomy of Section~\ref{sec:electrophysiology}:
\begin{equation}
\begin{aligned}
    \Zdepol &= \fdepol(\Zone) + \Udepol, \\
    \Zrepol &= \frepol(\Zone) + \Urepol,
\end{aligned}
\end{equation}
where additive terms $\Udepol$, $\Urepol$ absorb individual-specific electrophysiological variation. Waveform components are then synthesized,
\begin{equation}
    \Zqrs = \fqrs(\Zdepol), \qquad \Zstt = \fstt(\Zrepol),
\end{equation}
and a deconvolutional decoder reconstructs the segment with an environment term $\Uenv$ capturing lead-specific factors such as electrode placement:
\begin{equation}
    \hatX = \gdec(\Zqrs, \Zstt, \Uenv).
\end{equation}
Each additive term is drawn from $\Usys$ by its own amortized inference network, namely the \emph{depolarization}, \emph{repolarization}, and \emph{environment inference networks}. Each network is a 15-tap convolution over $\Usys$ whose 128 output channels split into mean and log-variance halves and are reparameterized as in Section~\ref{sec:backbone}. The decoder $\gdec$ has three parts: a \emph{decoder input projection}, a $1\times1$ convolution that collapses the 320-channel concatenation of $\Zqrs$, $\Zstt$, and $\Uenv$ to 256 channels; an \emph{upsampling stack} of four $\times2$ upsampling--convolution blocks (256$\rightarrow$128$\rightarrow$64$\rightarrow$32$\rightarrow$12 channels); and a final linear interpolation to the input length.

The reconstruction pathway serves two purposes. First, it is a fidelity check: the latent space must retain the waveform morphology that the ontology says matters, which Section~\ref{sec:representation} verifies quantitatively. Second, it is where interventions are applied: perturbing a subspace or sub-layer and re-decoding yields an inspectable change in the reconstructed waveform, the basis of the perturbation analyses in Section~\ref{sec:representation}.

\subsection{Training objective}
\label{sec:objective}

TRACE is trained by minimizing
\begin{equation}
    \Ltotal = \Lrecon + \lambdaKL \Lkl + \lambdatask \Ltask + \lambdaanchor \Ltopo,
\end{equation}
where $\Ltopo = \Rtopo$. The reconstruction term combines mean-squared and mean-absolute error with a physiological mask that gives extra weight to QRS and ST--T regions, so that the decoder is held responsible for the waveform evidence of Table~\ref{tab:ontology}. Its weight decays from 0.5 to 0.1 over training as the task shifts from representation fidelity to diagnosis. The KL term regularizes $\Usys$ toward $\N(0,\mathbf{I})$ with a warm start: the weight is held at 0 until epoch 10 and then ramps linearly to 0.001 over the following ten epochs, which prevents early over-regularization. The task term is binary cross-entropy over the isolated heads. The penalty weight was taken from the sensitivity sweep of Section~\ref{sec:ablations}. Section~\ref{sec:setup} gives the full training configuration.

\begin{algorithm}[!ht]
\caption{TRACE training. The ontology in Section~\ref{sec:ontology} fixes the subspace partition and the routing function $r$ before training; nothing in the loop below changes them.}
\label{alg:training}
\begin{algorithmic}[1]
\REQUIRE Dataset $\mathcal{D}$; routing $r$ and subspace split from Table~\ref{tab:ontology}; hyperparameters $\lambdaKL, \lambdatask, \lambdaanchor$; epochs $E$.
\ENSURE Trained parameters $\{\Phi_\theta, W_{z_1}, b_{z_1}, \fdepol, \frepol, \fqrs, \fstt, \gdec, \psi_k\}$.
\STATE Initialize all parameters.
\FOR{$e = 1$ to $E$}
    \STATE Set $\lambda_{\mathrm{recon}}$ by the schedule (0.5 decaying to 0.1); set $\lambdaKL$ by warm start ($0$ until epoch 10, then 0.001).
    \FOR{each mini-batch $(X, Y) \subset \mathcal{D}$}
        \STATE $h \leftarrow \Phi_\theta(X)$
        \STATE $\Usys \sim \N(\mu_\theta(h), \sigma_\theta^2(h))$ \COMMENT{reparameterized}
        \STATE $\Zone \leftarrow \sigma(W_{z_1}\Usys + b_{z_1})$; split into $[\Zperf;\Zstruc;\Zcond]$ per the ontology
        \STATE $\Zdepol \leftarrow \fdepol(\Zone) + \Udepol$; \quad $\Zrepol \leftarrow \frepol(\Zone) + \Urepol$
        \STATE $\Zqrs \leftarrow \fqrs(\Zdepol)$; \quad $\Zstt \leftarrow \fstt(\Zrepol)$
        \STATE $\hatX \leftarrow \gdec(\Zqrs, \Zstt, \Uenv)$
        \STATE $\hat{y}_k \leftarrow \psi_k(Z_{1,r(k)})$ for every category $k$ \COMMENT{hard routing}
        \STATE $\mathcal{L} \leftarrow \Lrecon + \lambdaKL \Lkl + \lambdatask \Ltask + \lambdaanchor \Rtopo$
        \STATE Update all parameters with AdamW on $\mathcal{L}$.
    \ENDFOR
\ENDFOR
\STATE \RETURN trained model.
\end{algorithmic}
\end{algorithm}

Because the routing and the partition are both fixed before training, the claims they carry are testable rather than assumed.

\subsection{Statistical rationale}\label{sec:theory}

This subsection derives the two statistical statements behind the training objective: the variational objective and a decorrelation bound.

\paragraph{Notation and factorization.} Let $Z=(Z_1,Z_2,Z_3)$ with $Z_2=(Z_{2,\mathrm{depol}},Z_{2,\mathrm{repol}})$ and $Z_3=(Z_{3,\mathrm{qrs}},Z_{3,\mathrm{stt}})$, and let $\Uenv$ denote the environment term of Section~\ref{sec:mechanism}. A generative factorization consistent with the main text is
\begin{multline}
p_\theta(X,\Usys,Z,\Uenv) = p(\Usys)\, p_\theta(Z_1\mid \Usys) \\
\quad \cdot\, p_\theta(Z_2\mid Z_1,\Usys)\, p_\theta(\Uenv\mid \Usys) \\
\quad \cdot\, p_\theta(Z_3\mid Z_2)\, p_\theta(X\mid Z_3,\Uenv),
\end{multline}
with amortized posterior
\begin{multline}
q_\phi(\Usys,Z,\Uenv\mid X) = q_\phi(\Usys\mid X)\, q_\phi(Z_1\mid \Usys) \\
\cdot\, q_\phi(Z_2\mid Z_1,\Usys)\, q_\phi(\Uenv\mid \Usys)\, q_\phi(Z_3\mid Z_2).
\end{multline}
In the implemented model these factors are the named modules of Sections~\ref{sec:backbone} and \ref{sec:mechanism}: the perception backbone and the system inference network implement $q_\phi(\Usys\mid X)$; the clinical-space projection implements $q_\phi(Z_1\mid\Usys)$; the pathway mappings with the depolarization, repolarization, and environment inference networks implement $q_\phi(Z_2\mid Z_1,\Usys)$ and $q_\phi(\Uenv\mid\Usys)$; and the QRS and ST--T synthesis mappings implement $q_\phi(Z_3\mid Z_2)$.

\paragraph{ELBO expansion.} Starting from the marginal likelihood and applying Jensen's inequality,
\begin{multline}
\log p_\theta(X) \geq \E_{q_\phi}\!\left[\log p_\theta(X,\Usys,Z,\Uenv)\right. \\
\left. - \log q_\phi(\Usys,Z,\Uenv\mid X)\right].
\end{multline}
Expanding terms,
\begin{multline}
\mathcal{L}_{\mathrm{ELBO}}(X) = \\
\quad \E_{q_\phi}\!\left[\log p_\theta(X\mid Z_3,\Uenv)\right] \\
\quad - D_{\mathrm{KL}}\!\left(q_\phi(\Usys\mid X)\,\|\,p(\Usys)\right) \\
\quad - \E_{q_\phi}\!\left[D_{\mathrm{KL}}\!\left(q_\phi(Z,\Uenv\mid \Usys,X)\right.\right. \\
\qquad \left.\left.\|\,p_\theta(Z,\Uenv\mid \Usys)\right)\right].
\end{multline}
In TRACE the intermediate transitions are deterministic mappings with additive noise injected in $\Udepol$, $\Urepol$, and $\Uenv$. The residual latent KL term is therefore absorbed into reconstruction mismatch and stochastic regularization, and the practical objective is
\begin{multline}
\mathcal{L}_{\mathrm{ELBO}}(X) \approx \E_{q_\phi}\!\left[\log p_\theta(X\mid Z_3,\Uenv)\right] \\
- D_{\mathrm{KL}}\!\left(q_\phi(\Usys\mid X)\,\|\,p(\Usys)\right).
\end{multline}
With $\Lrecon = -\E_{q_\phi}[\log p_\theta(X\mid Z_3,\Uenv)]$ and $\Lkl = D_{\mathrm{KL}}(q_\phi(\Usys\mid X)\,\|\,p(\Usys))$, minimizing $\Lrecon + \Lkl$ maximizes this bound. For weighted KL ($\lambdaKL \neq 1$), the objective is the Lagrangian of the constrained problem $\max \E_{q_\phi}[\log p_\theta(X\mid Z_3,\Uenv)]$ subject to $D_{\mathrm{KL}}(q_\phi(\Usys\mid X)\,\|\,p(\Usys)) \leq C$.

\paragraph{Why the orthogonality penalty bounds mutual information.} Consider two centered, whitened latent blocks $Z_i$, $Z_j$ with cross-covariance $C_{ij}=\mathrm{Cov}(Z_i,Z_j)$. Under a joint Gaussian approximation,
\begin{equation}
I(Z_i;Z_j) = -\tfrac{1}{2}\log\det\!\left(I - C_{ij}C_{ij}^{\top}\right).
\end{equation}
For $\lVert C_{ij}\rVert_2 < 1$,
\begin{equation}
I(Z_i;Z_j) \leq \frac{1}{2}\cdot\frac{\lVert C_{ij}\rVert_F^2}{1-\lVert C_{ij}\rVert_2^2} = \mathcal{O}(\lVert C_{ij}\rVert_F^2),
\end{equation}
and, for weak dependence, $I(Z_i;Z_j) = \tfrac{1}{2}\lVert C_{ij}\rVert_F^2 + o(\lVert C_{ij}\rVert_F^2)$. The empirical penalty $\Rtopo$ is a sample-level proxy for the squared cross-covariance, so the derivation above turns the penalty from a heuristic into a decorrelation objective with a mutual-information interpretation. The bound itself applies under the stated centering, whitening, and Gaussian assumptions, which the sigmoid latent $\Zone$ satisfies approximately. This is a mini-batch decorrelation statement, with the interpretation of Section~\ref{sec:orthogonality}.


\section{Experiments}\label{sec:experiments}

\subsection{Experimental setup}
\label{sec:setup}

\begin{figure*}[!ht]
    \centering
    \includegraphics[width=\textwidth]{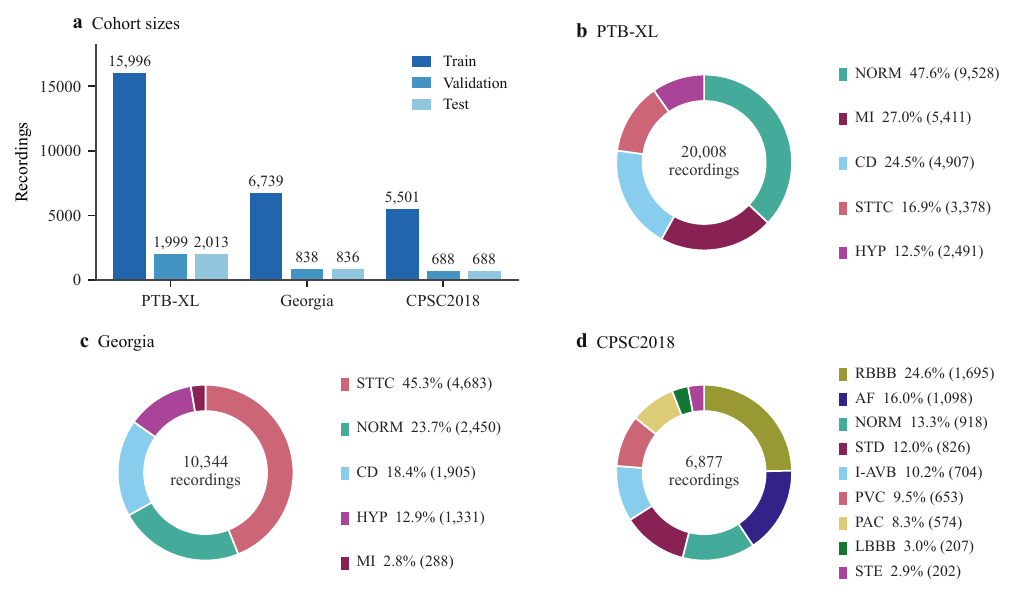}
    \caption{Dataset overview. (a) Training, validation, and test sizes. (b) PTB-XL superclass prevalence over the 20{,}008 label-filtered recordings used in this study. (c) Georgia label prevalence for the full 10,344-recording cohort, against the 8,413 used. (d) CPSC2018 label distribution over the full 6{,}877-recording cohort. Panels (b)--(c) report the percent of recordings positive under multi-label annotation, so prevalences may sum above 100\%; panel (d) bars sum to 100\%.}
    \label{fig:datasets}
\end{figure*}

\textbf{Datasets.} We evaluate on three 12-lead ECG datasets from PhysioNet \cite{goldberger2000physiobank}. PTB-XL \cite{wagner2020ptbxl,wagner2020ptbxlphysionet} is a large clinical cohort with the five-category ontology of Table~\ref{tab:ontology}. We retain the 20{,}008 recordings that carry a superclass label and use the standard fold split: folds 1--8 for training (15{,}996), fold 9 for validation (1{,}999), and fold 10 for testing (2{,}013). Georgia \cite{perezalday2020challenge,perezalday2022challengephysionet} provides cross-cohort evaluation under the same ontology, with stratified 80/10/10 splits over the 8{,}413 of its 10{,}344 recordings whose SNOMED-CT annotations map onto a superclass. CPSC2018 \cite{liu2018open} provides a nine-label setting with no structural class over all 6{,}877 of its recordings, in which the structure subspace is replaced by the rhythm subspace $\Zrhythm$. Figure~\ref{fig:datasets} summarizes the three cohorts.

\textbf{Baselines.} Table~\ref{tab:params} lists every model in the three groups used throughout this section: unconstrained classifiers, foundation models, and interpretable models. Unconstrained classifiers: FCN \cite{wang2017time}, ResNet1D, Inception1D \cite{ismail2020inceptiontime}, Transformer1D \cite{vaswani2017attention}. Self-supervised: MAE \cite{he2022masked} and SimCLR \cite{chen2020simple}. Interpretable: ConceptBottleneck \cite{koh2020concept}, trained here with supervised intermediate variables drawn from the same clinical vocabulary, namely the 48 SCP sub-category codes on PTB-XL, the 38 SNOMED codes on Georgia, and the ontology's category indicators on CPSC2018, where concepts and labels coincide. The comparison with TRACE therefore holds the knowledge source constant. Open-weight ECG foundation models, evaluated by linear probe on frozen features: ECGFounder \cite{ECGFounder2025}, pretrained on over ten million recordings across 150 classes, and HeartLang \cite{HeartLang2025}, pretrained on 800{,}035 MIMIC-IV recordings. Both use the same preprocessing as TRACE and logistic-regression probes. ST-MEM releases code but no pretrained weights.

\textbf{Training recipe.} Every model trained from scratch uses AdamW with weight decay $10^{-4}$, 30 epochs, and three seeds (42, 2025, 2026), reported as mean $\pm$ standard deviation. TRACE trains at a learning rate of $2\times10^{-4}$ with batch size 64 on every cohort; the baselines train at $10^{-3}$, at batch size 64 on PTB-XL and 128 on Georgia and CPSC2018, with ConceptBottleneck at 64 throughout and the self-supervised pretraining stages at 128. The protocol follows current reporting guidance for clinical prediction models \cite{collins2024tripodai} and the methodological recommendations for machine learning in medicine \cite{varoquaux2022machine}.

\textbf{Preprocessing.} Preprocessing is shared by every model within a cohort. PTB-XL and Georgia keep the first 10 seconds of each record, zero-padding shorter ones at the end, then apply a fourth-order Butterworth high-pass at $0.5$\,Hz with forward-backward filtering, and finally subtract the per-lead mean and divide by the per-lead standard deviation. CPSC2018 filters and normalizes first and then adjusts the length, with a random 10-second crop during training and a center crop at validation and test; random cropping is the only augmentation and applies to that cohort alone. Georgia and CPSC2018 signals are stored in acquisition units and rescaled to millivolts in the loader before filtering, using a cohort-specific raw-amplitude threshold (200 for Georgia and 20 for CPSC2018).

\textbf{Physiological mask.} The reconstruction term carries a mask that weights the loss 5.0 over the $\pm0.06$\,s QRS window around each R-peak and 2.0 over the P and T windows, with R-peaks detected on lead II at 0.3 of its peak absolute amplitude and a minimum spacing of $0.4$\,s. The mask applies to the reconstruction term alone and leaves the diagnostic path untouched.

\textbf{Implementation.} PyTorch 2.5.1, Python 3.12, Ubuntu 22.04, CUDA 12.4, on one NVIDIA RTX 4090 (24 GB).

\textbf{Parameter accounting.} TRACE stores 3.93M parameters, more than the standard classifiers because of its reconstruction pathway and hierarchical latent layers, and a factor of eight to eleven fewer than the 30.89M of ECGFounder and 44.56M of HeartLang. The MAE and SimCLR entries carry their self-supervised pretraining heads, and TRACE's entry carries its reconstruction pathway, the waveform-synthesis mappings and decoder that serve the fidelity and intervention analyses of Section~\ref{sec:representation}. The parameters that produce a diagnosis are the perception backbone, the system inference network, the clinical-space projection, and the isolated heads: 1.87M for TRACE and 0.95M for MAE and SimCLR. TRACE's reconstruction branch accounts for the other 2.06M and leaves every prediction unchanged, so it can be dropped at deployment.

\textbf{Foundation-model evaluation.} ECGFounder enters with its official 12-lead checkpoint and HeartLang through its pretrained feature extractor, both as logistic-regression probes on frozen features; the ECGFounder probe reads the 1024-dimensional penultimate representation rather than the 150-class output layer. All methods are scored on the same partition of each cohort. The two foundation probes are restricted to records that carry a superclass label, which leaves 2{,}000 and 2{,}001 of PTB-XL's 2{,}013 test recordings. Both foundation probes are deterministic: the feature extractor runs frozen in evaluation mode without augmentation, and the probe solver is not stochastic, so their scores are reported as single values.

\begin{table}[!ht]
\caption{Model sizes in millions of parameters.}
\label{tab:params}
\begin{tabular*}{\tblwidth}{@{}Lr@{}}
\toprule
Model & Parameters (M) \\
\midrule
\multicolumn{2}{l}{\textbf{Unconstrained classifiers}} \\
FCN & 0.28 \\
ResNet1D & 0.51 \\
Inception1D & 0.19 \\
Transformer1D & 0.90 \\
MAE & 1.14 \\
SimCLR & 0.98 \\
\midrule
\multicolumn{2}{l}{\textbf{Foundation models}} \\
ECGFounder & 30.89 \\
HeartLang & 44.56 \\
\midrule
\multicolumn{2}{l}{\textbf{Interpretable models}} \\
ConceptBottleneck & 0.52 \\
TRACE & 3.93 \\
\bottomrule
\end{tabular*}
\end{table}

\subsection{Main diagnostic performance}
\label{sec:main-results}

Table~\ref{tab:main} reports macro AUC and macro F1 across the three datasets. Both metrics appear because the ranking of multi-label methods is metric-dependent \cite{bogatinovski2022comprehensive}. Three findings stand out. First, TRACE outperforms ConceptBottleneck on every dataset and metric, on Georgia by 9.52 macro-F1 points (67.40 vs.\ 57.88), which shows that specified structure need not trade away accuracy. Second, TRACE exceeds the unconstrained classifiers on PTB-XL and Georgia while providing routing-level auditability they lack. On CPSC2018, TRACE reaches 64.54 macro F1 and 89.34 macro AUC. Third, TRACE trains from scratch on far fewer recordings than ECGFounder's ten-million-scale pretraining corpus. It is ahead of that model on PTB-XL by 0.54 AUC points (91.97 vs.\ 91.43) and on Georgia by 0.93 (87.31 vs.\ 86.38), at roughly an eighth of the parameter count. ECGFounder, by contrast, exposes no decision pathway through its linear probe. HeartLang transfers less readily across all three cohorts.

\begin{table*}[!ht]
\caption{Macro AUC and macro F1 across PTB-XL, Georgia, and CPSC2018 in percent, with the best value in bold and the second best underlined. Models trained under the shared recipe are reported as mean $\pm$ std over three seeds. The two foundation models are frozen-feature linear probes with a deterministic solver, so they are reported as single values. Every method is scored on the same partition of each cohort.}
\label{tab:main}
\begin{tabular*}{\tblwidth}{@{}lLLLLLL@{}}
\toprule
Method & \multicolumn{2}{c}{PTB-XL} & \multicolumn{2}{c}{Georgia} & \multicolumn{2}{c}{CPSC2018} \\
\cmidrule(lr){2-3} \cmidrule(lr){4-5} \cmidrule(lr){6-7}
 & Macro AUC & Macro F1 & Macro AUC & Macro F1 & Macro AUC & Macro F1 \\
\midrule
FCN & 91.13 $\pm$ 0.09 & 71.88 $\pm$ 0.43 & 82.10 $\pm$ 0.25 & 57.35 $\pm$ 0.97 & 90.05 $\pm$ 0.32 & 61.96 $\pm$ 1.01 \\
ResNet1D & 91.41 $\pm$ 0.15 & \underline{72.22} $\pm$ 0.40 & 81.73 $\pm$ 0.70 & 56.34 $\pm$ 0.45 & 91.55 $\pm$ 0.19 & 64.64 $\pm$ 0.50 \\
Inception1D & \underline{91.81} $\pm$ 0.08 & 71.89 $\pm$ 0.11 & 84.73 $\pm$ 0.33 & 61.04 $\pm$ 1.36 & \underline{93.52} $\pm$ 0.30 & \underline{70.46} $\pm$ 0.20 \\
Transformer1D & 90.20 $\pm$ 0.19 & 70.25 $\pm$ 0.23 & 80.76 $\pm$ 0.30 & 55.68 $\pm$ 1.03 & 90.24 $\pm$ 0.72 & 62.48 $\pm$ 1.38 \\
MAE & 89.43 $\pm$ 0.25 & 68.90 $\pm$ 0.46 & 81.21 $\pm$ 1.50 & 56.02 $\pm$ 2.06 & 89.76 $\pm$ 0.59 & 63.01 $\pm$ 1.16 \\
SimCLR & 88.79 $\pm$ 0.09 & 67.69 $\pm$ 0.47 & 82.39 $\pm$ 0.60 & 56.59 $\pm$ 1.76 & 92.54 $\pm$ 0.16 & 67.04 $\pm$ 1.17 \\
\midrule
ECGFounder & 91.43 & 71.23 & \underline{86.38} & \underline{64.60} & \textbf{95.75} & \textbf{77.45} \\
HeartLang & 89.58 & 68.76 & 80.74 & 58.52 & 88.36 & 59.21 \\
\midrule
ConceptBottleneck & 90.89 $\pm$ 0.18 & 72.02 $\pm$ 0.13 & 81.07 $\pm$ 0.39 & 57.88 $\pm$ 1.25 & 86.80 $\pm$ 0.63 & 59.79 $\pm$ 0.05 \\
TRACE & \textbf{91.97} $\pm$ 0.05 & \textbf{72.59} $\pm$ 0.13 & \textbf{87.31} $\pm$ 0.09 & \textbf{67.40} $\pm$ 1.60 & 89.34 $\pm$ 0.43 & 64.54 $\pm$ 2.41 \\
\bottomrule
\end{tabular*}
\end{table*}

The CPSC2018 result has a structural explanation. Its label set is larger and has no structural class, and the routing of Table~\ref{tab:routing_app} packs it more tightly than the five-category routing packs PTB-XL. Three labels (AF, PAC, PVC) share the 8-dimensional rhythm subspace, and three more (I-AVB, LBBB, RBBB) share the 8-dimensional conduction subspace; PTB-XL and Georgia place a single label in each. Section~\ref{sec:ablations} quantifies that constraint. Reallocating the same 32 dimensions on CPSC2018 to 8/12/12 drops macro F1 from 64.54 to 18.87, against a 7.6-point F1 cost for the equivalent 16--4--4 reallocation on PTB-XL. Swapping subspace definitions and the routing table leaves the training framework untouched, and Table~\ref{tab:routing_app} reports the adapted routing for CPSC2018.

\begin{figure*}[!ht]
    \centering
    \includegraphics[width=\textwidth]{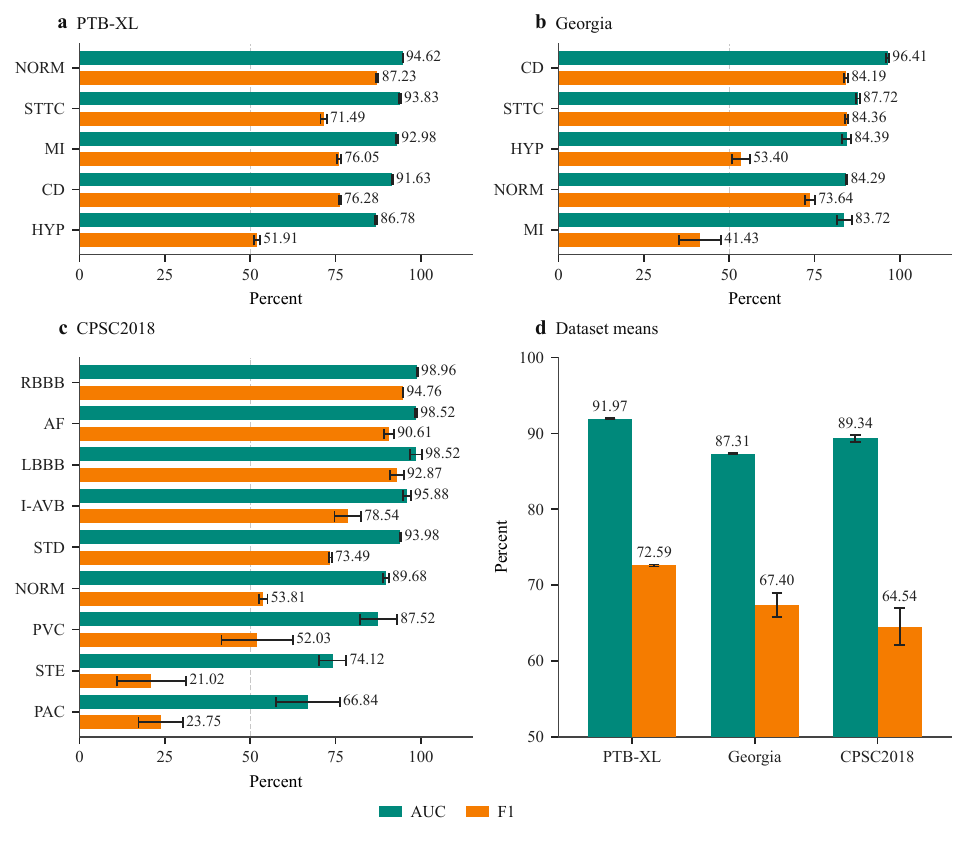}
    \caption{Class-wise discrimination and thresholding for TRACE. In (a)--(c) each class shows its AUC in teal and its F1 in orange as paired horizontal bars; values are percent, mean $\pm$ std over three seeds, with whiskers showing the seed standard deviation. Classes are ordered by descending AUC, and a dashed line marks the 50\% chance level. (a) PTB-XL fold-10 test set, (b) Georgia cross-cohort test split, (c) CPSC2018 with per-class optimal thresholds across nine labels. (d) Per-dataset macro AUC and macro F1 as vertical bars.}
    \label{fig:classwise}
\end{figure*}

Figure~\ref{fig:classwise} reports per-class AUC and F1 under the same optimal-threshold protocol. The labels that route through their designated subspace track their pathway well, and rare classes track their prevalence, as expected under per-class thresholding.

On CPSC2018 (Figure~\ref{fig:classwise}(c)), PAC, PVC, and STE show high variance and threshold sensitivity: PAC and PVC appear sporadically within a 10-second window, and STE has 24 test samples.

\subsection{Routing-contract verification}
\label{sec:routing-verification}

We verify the routing contract of Section~\ref{sec:method} with two analyses read jointly, following the amnesic probing method \cite{elazar2021amnesic,belinkov2022probing}. The contract makes two empirical claims: each subspace contains its intended label information, and each head consumes only its designated subspace. Linear probing tests the first claim, observational alignment, and concept erasure tests the second, what each head causally used. The two tests answer different questions: a probe finds information the head may ignore, while erasure does not say what the representation contains. Because hard routing gives each head a disjoint input slice, the wiring is checkable mechanically, and encoder cross-subspace information is measured by the probe matrix.

\begin{figure*}[!ht]
    \centering
    \includegraphics[width=\textwidth]{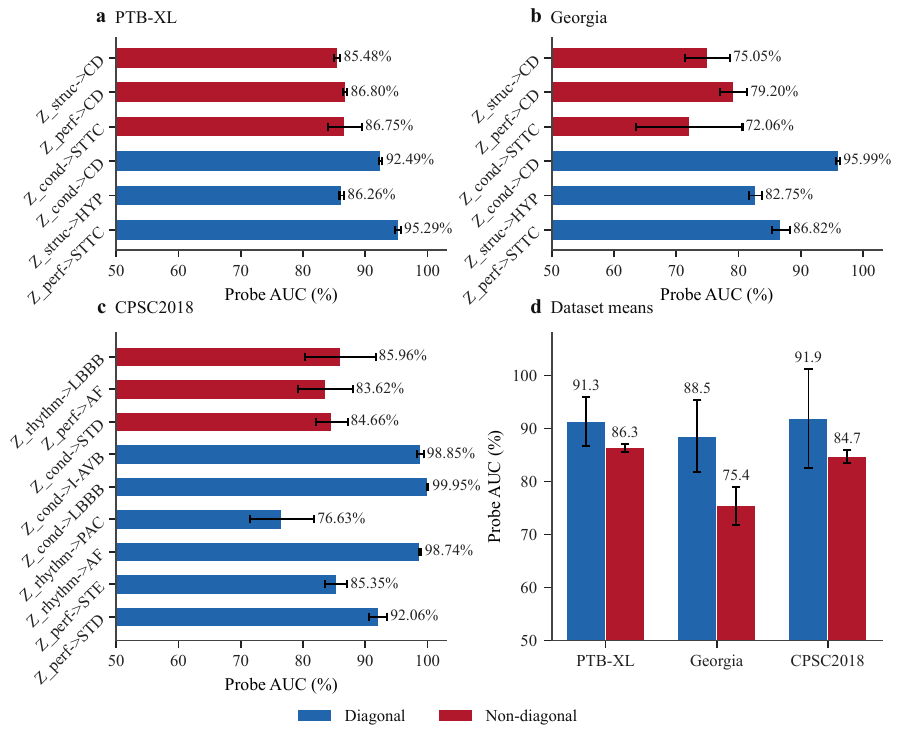}
    \caption{Linear probe AUC in percent, mean $\pm$ std over three seeds. Blue: the intended diagonal subspace-label pairs; red: the off-diagonal cross-subspace pairs. (a) PTB-XL and (b) Georgia under the perfusion/structure/conduction routing; (c) CPSC2018 under the adapted perfusion/rhythm/conduction routing; (d) per-dataset means of the diagonal and off-diagonal probes, averaged over the pairs plotted in (a)--(c), with error bars showing the across-pair standard deviation. Each subspace in (a)--(c) is probed against its designated label and against the labels of the other subspaces; where a diagonal label has no cross-subspace counterpart (HYP on PTB-XL and Georgia; STE, PAC and I-AVB on CPSC2018) it is plotted alone.}
    \label{fig:probe}
\end{figure*}

\textbf{Probes.} Figure~\ref{fig:probe} shows the routing-probe picture, each measured probe plotted against its designated route: $Z_{\mathrm{perf}}\!\rightarrow\!\mathrm{STTC}$ reached $95.29\pm0.47\%$ AUC on PTB-XL and $Z_{\mathrm{cond}}\!\rightarrow\!\mathrm{CD}$ reached $95.99\pm0.31\%$ on Georgia (mean $\pm$ std over three seeds); the remaining diagonal pair $Z_{\mathrm{perf}}\!\rightarrow\!\mathrm{MI}$ probed at $93.92\%$ on PTB-XL and $82.42\%$ on Georgia, three-seed means; on CPSC2018 (Figure~\ref{fig:probe}(c)) $Z_{\mathrm{rhythm}}\!\rightarrow\!\mathrm{AF}$ reached $98.74\pm0.17\%$ and $Z_{\mathrm{cond}}\!\rightarrow\!\mathrm{LBBB}$ $99.95\pm0.07\%$. Off-diagonal probes were non-zero, with $Z_{\mathrm{cond}}\!\rightarrow\!\mathrm{STTC}$ at $86.75\pm2.75\%$ on PTB-XL. Holding CD status fixed does not remove this association ($81.95\pm2.29\%$ within CD-positive and $84.35\pm1.80\%$ within CD-negative recordings), and CD and STTC are close to independent in this cohort ($15.3\%$ versus $16.8\%$ STTC prevalence), so the residual association is representational rather than an artifact of clinical co-morbidity. An association observed in the representation need not be one a head used, and the erasure analysis separates the two. The diagonal advantage is a property of the design: the anchor alignment and head gradients push each subspace toward its label, so a high diagonal probe confirms that the routing contract was enforced. The probes establish the off-diagonal structure and the diagonal margin; representation-level semantics rest on the mechanism perturbations of Section~\ref{sec:representation}.

\textbf{Erasure.} Table~\ref{tab:erasure} reports targeted erasure on all three datasets, with three-seed mean $\pm$ std in the baseline column. On PTB-XL, zeroing $Z_{\mathrm{perf}}$ collapsed MI and STTC to chance (50\%) while CD stayed at its $91.63\pm0.11$ baseline. Zeroing $Z_{\mathrm{cond}}$ collapsed CD while STTC and MI remained at $93.83\pm0.26$ and $92.98\pm0.31$. Non-target heads are unchanged because they read a disjoint slice, as the construction requires, and the cross-subspace information that remains in the encoder is what the probe matrix of Figure~\ref{fig:probe} measures. For every subspace-label pair with an off-diagonal counterpart, the diagonal probe exceeded the off-diagonal probe on three-seed means, within both CD strata as well. Read jointly, each subspace was most informative for its intended condition (probe), and the wiring that confines each head to its own slice holds by construction (erasure).

\begin{table}[!ht]
\caption{Targeted erasure AUC in percent, with the baseline column as mean $\pm$ std over three seeds. Entries in bold are the head whose own subspace was erased, collapsed to the chance level of 50.00. NORM reads the unmodified full $\Zone$, so it is invariant under every erasure by construction; CPSC2018 uses $\Zrhythm$ in place of $\Zstruc$.}
\label{tab:erasure}
{\setlength{\tabcolsep}{3pt}
\begin{tabular*}{\tblwidth}{@{}Lcccc@{}}
\toprule
\multicolumn{5}{l}{\textbf{PTB-XL}} \\
\cmidrule(lr){1-5}
Label & Baseline & w/o $\Zperf$ & w/o $\Zstruc$ & w/o $\Zcond$ \\
\midrule
NORM & $94.62 \pm 0.02$ & 94.62 & 94.62 & 94.62 \\
MI & $92.98 \pm 0.31$ & \textbf{50.00} & 92.98 & 92.98 \\
STTC & $93.83 \pm 0.26$ & \textbf{50.00} & 93.83 & 93.83 \\
CD & $91.63 \pm 0.11$ & 91.63 & 91.63 & \textbf{50.00} \\
HYP & $86.78 \pm 0.31$ & 86.78 & \textbf{50.00} & 86.78 \\
\midrule
\multicolumn{5}{l}{\textbf{Georgia}} \\
\cmidrule(lr){1-5}
Label & Baseline & w/o $\Zperf$ & w/o $\Zstruc$ & w/o $\Zcond$ \\
\midrule
NORM & $84.29 \pm 0.24$ & 84.29 & 84.29 & 84.29 \\
MI & $83.72 \pm 2.19$ & \textbf{50.00} & 83.72 & 83.72 \\
STTC & $87.72 \pm 0.47$ & \textbf{50.00} & 87.72 & 87.72 \\
CD & $96.41 \pm 0.41$ & 96.41 & 96.41 & \textbf{50.00} \\
HYP & $84.39 \pm 1.25$ & 84.39 & \textbf{50.00} & 84.39 \\
\midrule
\multicolumn{5}{l}{\textbf{CPSC2018}} \\
\cmidrule(lr){1-5}
Label & Baseline & w/o $\Zperf$ & w/o $\Zrhythm$ & w/o $\Zcond$ \\
\midrule
NORM & $89.68 \pm 0.87$ & 89.68 & 89.68 & 89.68 \\
AF & $98.52 \pm 0.30$ & 98.52 & \textbf{50.00} & 98.52 \\
I-AVB & $95.88 \pm 1.09$ & 95.88 & 95.88 & \textbf{50.00} \\
LBBB & $98.52 \pm 1.75$ & 98.52 & 98.52 & \textbf{50.00} \\
RBBB & $98.96 \pm 0.11$ & 98.96 & 98.96 & \textbf{50.00} \\
PAC & $66.84 \pm 9.37$ & 66.84 & \textbf{50.00} & 66.84 \\
PVC & $87.52 \pm 5.44$ & 87.52 & \textbf{50.00} & 87.52 \\
STD & $93.98 \pm 0.21$ & \textbf{50.00} & 93.98 & 93.98 \\
STE & $74.12 \pm 3.98$ & \textbf{50.00} & 74.12 & 74.12 \\
\bottomrule
\end{tabular*}}
\end{table}

\subsection{Ablations}
\label{sec:ablations}

\begin{figure*}[!ht]
    \centering
    \includegraphics[width=\textwidth]{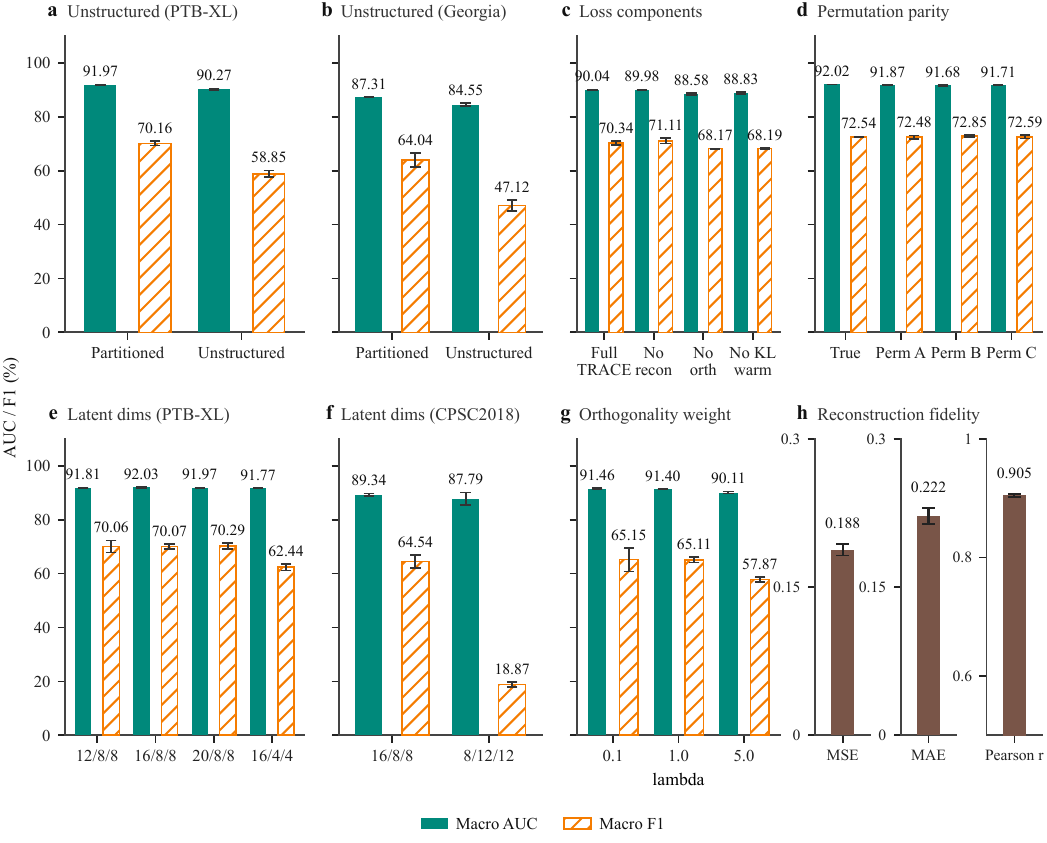}
    \caption{Ablations. Solid bars show macro AUC and hatched bars macro F1, in percent, mean $\pm$ std over three seeds, with whiskers for the seed standard deviation. (a)--(b) Structured vs.\ unstructured latent on PTB-XL and Georgia, both sides evaluated at a fixed 0.5 decision threshold. (c) Loss-component ablation on PTB-XL under the fold-10 optimal-threshold protocol. (d) Capacity-matched permutation control on PTB-XL: the true routing and the three permuted assignments of the four blocks. (e)--(f) Latent-dimension sensitivity on PTB-XL and CPSC2018. (g) Orthogonality-weight sweep. (h) Reconstruction fidelity on PTB-XL test: MSE, MAE, and Pearson $r$, each on its own scale.}
    \label{fig:ablations}
\end{figure*}

\textbf{Unstructured latent baseline.} A TRACE variant whose 32-dimensional latent space is not partitioned is trained with all heads reading the full $\Zone$ and the orthogonality penalty removed, with encoder, decoder, KL regularization, and the remaining hyperparameters identical; the PTB-XL unstructured run used batch size 256 instead of 64, and the Georgia unstructured run trained with automatic mixed precision enabled. Both variants are measured under the same fixed-0.5 protocol over three seeds. Figure~\ref{fig:ablations}(a)--(b) show the consequence. Removing the partition cost 1.70 AUC and 11.30 macro-F1 points on PTB-XL, and 2.76 AUC and 16.92 macro-F1 points on Georgia. The partition matters on both metrics and in both cohorts under this fixed-head design.

\textbf{Component-level ablation.} Figure~\ref{fig:ablations}(c) isolates each loss term on PTB-XL under the optimal-threshold fold-10 protocol. Removing the orthogonality penalty cost 1.5 AUC and 2.2 F1 points; removing reconstruction cost no AUC but widened F1 variance across seeds; removing the KL warm start cost 2.2 F1 points. Each component contributed, and the orthogonality penalty added to what hard routing alone provides.

\textbf{Dimension and regularization sensitivity.} Perfusion dimension 12/16/20 with structure and conduction fixed at 8 each left AUC within 0.22 points, and shrinking structure and conduction to 4 each (16--4--4) cost 7.6 F1 points, which places a minimum per-subspace capacity. The orthogonality weight was stable for $\lambdaanchor \in [0.1, 1.0]$, at AUC 91.46 against 91.40. At $\lambdaanchor = 5.0$ over-regularization dropped AUC to 90.11 and F1 to 57.87. Figure~\ref{fig:ablations}(e)--(f) and (g) give the full sweeps.

\textbf{Capacity-matched permutation control.} Permuting the subspace-head routing and retraining tests whether the \emph{specific} ontology matters beyond any fixed partition of equal capacity. A naive permutation reallocates head capacity together with routing, which confounds the two factors, so the control is capacity-matched: $\Zone$ is partitioned into four 8-dimensional blocks that are permuted as a whole, every diagnosis head receives an 8-dimensional input in every assignment, and the auxiliary anchor alignment follows the assigned routing. The four blocks, in the order the panel labels them, are the two halves of $\Zperf$ (Perf-A, Perf-B), $\Zstruc$ (Struc), and $\Zcond$ (Cond). Writing each assignment as the blocks that MI, STTC, CD, and HYP receive in that order, the true routing is Perf-A/Perf-B/Cond/Struc, and the three permutations are Struc/Cond/Perf-A/Perf-B (A), Cond/Struc/Perf-B/Perf-A (B), and Perf-B/Perf-A/Struc/Cond (C). Figure~\ref{fig:ablations}(d) reports all four assignments under the same fold-10 optimal-threshold protocol as the component ablation above.

All four assignments land within 0.34 AUC points of one another. Pooling the nine seed-matched permuted runs, the true routing leads the permutations by 0.26 AUC points (paired $t(8)=4.89$, $p=0.001$) and is level on macro F1 at $-0.10$ points (paired $t(8)=-0.52$, $p=0.619$), at the resolution of three seeds per assignment. Under the component-ablation protocol the same seeds reproduce the main-table numbers exactly (91.97/72.59), which confirms protocol consistency. Two readings follow. First, the ontology carries no routing-specific shortcut: arbitrary permutations of equal capacity come within a third of an AUC point of the ontology routing. Second, the specified routing gains a small but significant ranking margin at no macro-F1 cost, so transparency and the ability to steer the model cost nothing in either metric. The ontology supplies the routing contract as the mechanism that makes this re-specifiable. Its correctness as a clinical mapping is a normative specification (Section~\ref{sec:preliminaries}); the permutation control shows that the naming contributes little to prediction.

\subsection{Representation-level evidence}
\label{sec:representation}

Routing-level transparency is a property of the architecture. This subsection reports the empirical question behind it: what the model does under intervention at the layer where the architecture's commitments are most directly inspectable.

\begin{figure*}[tbp]
    \centering
    \includegraphics[width=\textwidth]{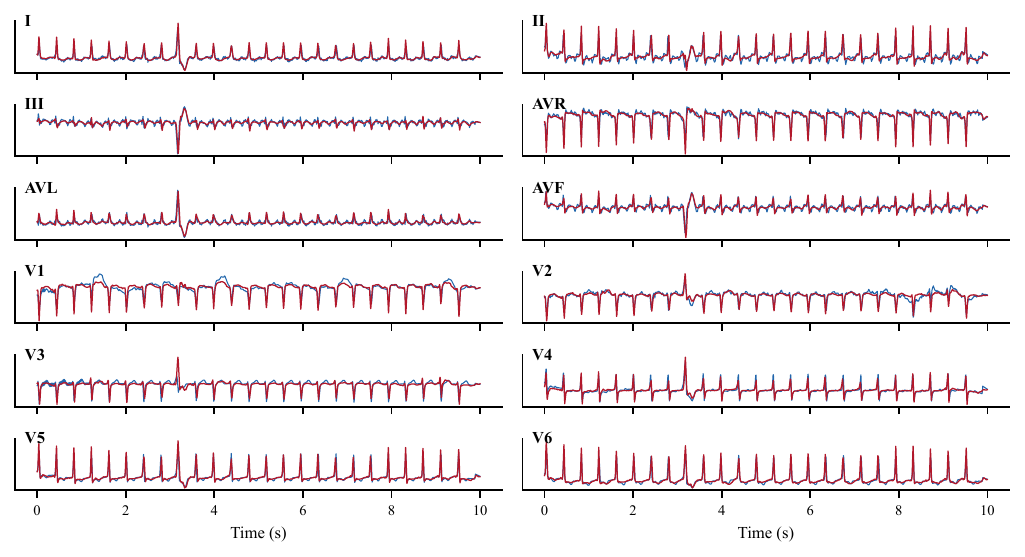}\\[-6pt]
    {\footnotesize\textbf{(a)} sample index 42}\\[2pt]
    \includegraphics[width=\textwidth]{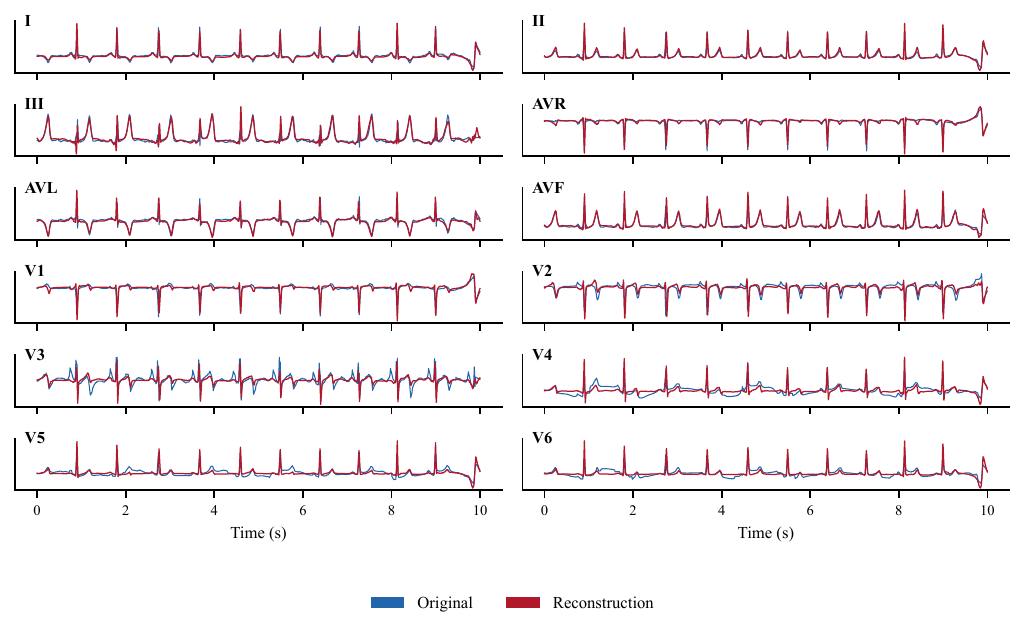}\\[-6pt]
    {\footnotesize\textbf{(b)} sample index 100}
    \vspace{-4pt}
    \caption{PTB-XL reconstruction overlays for two test recordings over the 10-second segment: original in blue and reconstruction in red for all 12 leads, two per row. (a) Sample index 42, (b) sample index 100.}
    \label{fig:reconstruction}
\end{figure*}

\textbf{Mechanism layer.} On all 2{,}013 test recordings over three seeds, with R-peak-anchored QRS and ST--T windows, zeroing the depolarization pathway and re-decoding changed QRS energy by $-0.9\pm3.1\%$ and ST--T energy by $-0.4\pm2.7\%$. Zeroing the repolarization pathway changed QRS by $+7.3\pm1.1\%$ and ST--T by $+7.7\pm1.3\%$ (mean $\pm$ std over seeds), with fixed-window estimates agreeing in magnitude and direction. At this granularity both interventions modulate both windows, so mechanism-layer intervention acts on the reconstructed waveform as a whole rather than selectively on individual waveform components. The same protocol applied to the three permuted routings of Figure~\ref{fig:ablations}(d) produced behavior of the same magnitude and direction, so the pathway structure itself sets what the mechanism decomposition does, not the subspace naming. The reconstruction pathway behind these measurements is itself accurate, with MSE $0.188\pm0.006$, MAE $0.222\pm0.008$, and Pearson $r=0.905\pm0.003$ (mean $\pm$ std over three seeds) between original and reconstructed 12-lead signals on PTB-XL test (Figure~\ref{fig:ablations}(h)). Figure~\ref{fig:reconstruction} overlays two recordings. The two branches are coupled pathways.

\subsection{Robustness under realistic perturbations}
\label{sec:robustness}

Table~\ref{tab:robustness} reports macro F1 under four conditions: clean input, additive Gaussian noise, baseline wander, and lead masking. Every perturbation is applied to the test input alone, after the shared preprocessing of Section~\ref{sec:setup} and without retraining or adaptation, and each targets one failure mode of the recording chain. Per-lead normalization fixes every lead to unit standard deviation, so the magnitudes below are multiples of a lead's signal standard deviation. Noise adds independent zero-mean Gaussian samples of standard deviation $0.2$ to all twelve leads, modeling contaminated electrode contact. Drift adds a $0.2$\,Hz sinusoid of unit amplitude, two cycles over the 10-second window, identically to all twelve leads, modeling respiration-induced wander. Its root-mean-square is $1/\sqrt{2}$ of a signal standard deviation, and the band is the one the $0.5$\,Hz high-pass removes, so the condition measures the diagnostic path with that stage bypassed. Mask sets two of the twelve leads, drawn uniformly at random for every recording, to zero, modeling disconnected electrodes. Per-class decision thresholds are tuned on the clean validation split and held fixed across the four conditions, so no method gains from re-tuning under corruption.

TRACE leads every model trained under the shared recipe on PTB-XL and Georgia under the clean, noise, and mask conditions. On those two cohorts, ECGFounder's pretraining advantage appears under baseline drift and under lead masking on Georgia (59.34 vs.\ 58.84); on CPSC2018 it leads in every condition. Drift is the one condition that reorders the shared-recipe models. On PTB-XL, TRACE holds at 52.09, ahead of all four standard classifiers, while the two self-supervised objectives hold above 56; on Georgia and CPSC2018 those models span 23--49 and 25--50 respectively. Because the recipe includes neither adversarial training nor any perturbation-specific augmentation, the resistance follows from the construction.

\begin{table*}[!ht]
\caption{Robustness across PTB-XL, Georgia, and CPSC2018: macro F1 in percent, best value in bold, second best underlined. Averaging as in Table~\ref{tab:main}, with decision thresholds fixed at their clean-validation values.}
\label{tab:robustness}
{\setlength{\tabcolsep}{4pt}
\begin{tabular*}{\tblwidth}{@{}Lcccc@{}}
\toprule
\multicolumn{5}{l}{\textbf{PTB-XL}} \\
\cmidrule(lr){1-5}
Method & Clean & Noise & Drift & Mask \\
\midrule
FCN & 71.88 $\pm$ 0.43 & 71.78 $\pm$ 0.22 & 27.56 $\pm$ 7.15 & 62.14 $\pm$ 0.89 \\
ResNet1D & \underline{72.22} $\pm$ 0.40 & \underline{71.80} $\pm$ 0.59 & 30.39 $\pm$ 7.80 & 63.13 $\pm$ 1.20 \\
Inception1D & 71.90 $\pm$ 0.12 & 70.87 $\pm$ 0.59 & 45.75 $\pm$ 10.28 & 63.52 $\pm$ 1.72 \\
Transformer1D & 70.25 $\pm$ 0.23 & 70.18 $\pm$ 0.39 & 45.97 $\pm$ 0.45 & 66.18 $\pm$ 0.75 \\
MAE & 68.90 $\pm$ 0.46 & 68.88 $\pm$ 0.52 & \underline{59.28} $\pm$ 1.89 & 66.69 $\pm$ 0.76 \\
SimCLR & 67.69 $\pm$ 0.47 & 67.66 $\pm$ 0.52 & 56.60 $\pm$ 2.03 & 63.61 $\pm$ 0.70 \\
\midrule
ECGFounder & 71.23 & 67.89 & \textbf{66.53} & \underline{67.19} \\
HeartLang & 68.76 & 67.63 & 21.83 & 55.24 \\
\midrule
ConceptBottleneck & 72.02 $\pm$ 0.13 & 71.48 $\pm$ 0.12 & 42.72 $\pm$ 3.40 & 63.52 $\pm$ 1.53 \\
TRACE & \textbf{72.60} $\pm$ 0.14 & \textbf{72.10} $\pm$ 0.54 & 52.09 $\pm$ 2.63 & \textbf{67.27} $\pm$ 0.45 \\
\bottomrule
\end{tabular*}

\vspace{1.6mm}

\begin{tabular*}{\tblwidth}{@{}Lcccc@{}}
\toprule
\multicolumn{5}{l}{\textbf{Georgia}} \\
\cmidrule(lr){1-5}
Method & Clean & Noise & Drift & Mask \\
\midrule
FCN & 57.35 $\pm$ 0.97 & 53.42 $\pm$ 1.50 & 28.48 $\pm$ 1.99 & 50.75 $\pm$ 1.47 \\
ResNet1D & 56.34 $\pm$ 0.45 & 54.26 $\pm$ 0.59 & 32.79 $\pm$ 5.11 & 52.57 $\pm$ 2.22 \\
Inception1D & 61.04 $\pm$ 1.36 & 47.93 $\pm$ 1.70 & 23.43 $\pm$ 3.85 & 53.20 $\pm$ 0.58 \\
Transformer1D & 55.68 $\pm$ 1.03 & 55.51 $\pm$ 1.29 & \underline{48.76} $\pm$ 1.92 & 53.48 $\pm$ 0.99 \\
MAE & 56.03 $\pm$ 2.07 & 55.72 $\pm$ 1.53 & 38.43 $\pm$ 1.12 & 53.45 $\pm$ 2.37 \\
SimCLR & 56.59 $\pm$ 1.76 & 56.59 $\pm$ 1.43 & 39.97 $\pm$ 6.51 & 54.68 $\pm$ 1.15 \\
\midrule
ECGFounder & \underline{64.60} & \underline{57.63} & \textbf{58.84} & \textbf{59.34} \\
HeartLang & 58.52 & 51.89 & 27.76 & 47.40 \\
\midrule
ConceptBottleneck & 57.88 $\pm$ 1.25 & 55.52 $\pm$ 0.55 & 36.02 $\pm$ 5.94 & 49.37 $\pm$ 1.19 \\
TRACE & \textbf{67.41} $\pm$ 1.60 & \textbf{63.80} $\pm$ 2.30 & 44.10 $\pm$ 3.63 & \underline{58.84} $\pm$ 1.90 \\
\bottomrule
\end{tabular*}

\vspace{1.6mm}

\begin{tabular*}{\tblwidth}{@{}Lcccc@{}}
\toprule
\multicolumn{5}{l}{\textbf{CPSC2018}} \\
\cmidrule(lr){1-5}
Method & Clean & Noise & Drift & Mask \\
\midrule
FCN & 61.96 $\pm$ 1.01 & 53.53 $\pm$ 0.81 & 24.59 $\pm$ 1.25 & 48.27 $\pm$ 1.10 \\
ResNet1D & 64.64 $\pm$ 0.50 & 63.57 $\pm$ 0.42 & 28.57 $\pm$ 7.07 & 51.16 $\pm$ 1.39 \\
Inception1D & \underline{70.44} $\pm$ 0.21 & 64.11 $\pm$ 1.56 & 27.55 $\pm$ 2.92 & 57.27 $\pm$ 1.77 \\
Transformer1D & 62.48 $\pm$ 1.37 & 62.63 $\pm$ 1.79 & \underline{50.16} $\pm$ 0.61 & 53.25 $\pm$ 1.42 \\
MAE & 63.01 $\pm$ 1.16 & 62.89 $\pm$ 1.43 & 39.27 $\pm$ 5.59 & 60.83 $\pm$ 1.26 \\
SimCLR & 67.04 $\pm$ 1.17 & \underline{67.11} $\pm$ 0.91 & 49.92 $\pm$ 1.92 & \underline{62.80} $\pm$ 1.26 \\
\midrule
ECGFounder & \textbf{77.45} & \textbf{69.58} & \textbf{71.52} & \textbf{70.89} \\
HeartLang & 59.21 & 56.67 & 21.80 & 43.55 \\
\midrule
ConceptBottleneck & 59.79 $\pm$ 0.06 & 58.42 $\pm$ 0.80 & 36.29 $\pm$ 6.51 & 48.93 $\pm$ 2.02 \\
TRACE & 64.51 $\pm$ 2.43 & 62.06 $\pm$ 2.80 & 42.03 $\pm$ 2.88 & 55.33 $\pm$ 0.23 \\
\bottomrule
\end{tabular*}}
\end{table*}


\section{Discussion}\label{sec:discussion}

\subsection{Specified versus learned latent structure}

ECG datasets carry diagnostic labels rather than annotations of the physiological mechanisms behind them. Under label-only supervision, learned structure has no route to the ontology of Table~\ref{tab:ontology}, and whatever structure emerges is unverifiable from the labels alone. Specifying the partition in advance sidesteps that: the routing is fixed by cardiological knowledge rather than inferred from the labels. The capacity-matched permutation control of Section~\ref{sec:ablations} isolates what the specification contributes: the named clinical pathway credited with the diagnosis, and not the thresholded performance, since re-specification leaves macro F1 level. The same isolation the routing enforces is visible under corruption. ConceptBottleneck shares one bottleneck across all its predictions and degrades on every cohort, whereas the isolated heads confine the damage to the pathway the corruption entered, which is consistent with the robustness margins of Section~\ref{sec:robustness}.

\subsection{Value of routing-level transparency}

Routing-level transparency means a clinician can be told \emph{which knowledge pathway} produced a diagnosis: an MI prediction comes from the perfusion pathway, by construction, every time. Its value is the \emph{auditability of the model's design choices}, which is a different property from per-patient explanation. When a prediction is wrong, the failure is localized to a named pathway with a named clinical meaning, so debugging, data collection, and ontology re-specification can all be aimed at that pathway. A monolithic classifier or a foundation-model probe offers no such localization. The per-patient question, why this recording and this severity, is addressed instead at the representation level of Section~\ref{sec:representation}.

\subsection{Causal claims and what observational corpora support}

Under retrospective, label-only corpora, the supportable statements are observational alignment (probes), architectural routing (erasure), and waveform modulation under intervention in the model itself (perturbation). A claim that the subspaces recover the true physiological mechanisms would need more than these corpora: interventional or paired supervision, such as recordings before and after revascularization or ablation, is what would ground a statement that a latent direction \emph{is} ischemia. The identifiability literature explains why. A causal representation and its structure cannot be recovered from observational data alone, and become identifiable only under weak supervision from paired samples before and after unknown interventions \cite{brehmer2022weakly}. Causal machine learning for clinical outcomes is likewise framed as a distinct problem from associational prediction, resting on counterfactual quantities and explicit assumptions \cite{feuerriegel2024causal}. Paired-supervision ECG collections would enable the move from pathway auditability to causally grounded explanations.

\subsection{Limitations}

Our evaluation is retrospective on three public corpora; prospective utility across sites, devices, and workflows remains to be established. The ontology is re-specified per label taxonomy, as done for CPSC2018, and multi-system conditions that span pathways enter the model as elevated probabilities across single-subspace heads. Calibration at clinical operating points remains open, a property distinct from discrimination that modern networks do not target by default \cite{vancalster2019calibration,guo2017calibration}. Distribution-free conformal frameworks are the natural next step, both for dynamic biological models \cite{portela2025conformal} and for general metric spaces \cite{lugosi2024uncertainty}, with time-series formulations that relax the exchangeability assumption \cite{xu2023conformal}. Reliability against inputs outside the learned distribution is a separate failure mode, addressed by out-of-distribution detection \cite{yu2025trustworthy}. Selective control of individual waveform components needs finer-grained supervision than these corpora provide, and per-patient feature-level semantics of individual latent dimensions is the open question that Section~\ref{sec:preliminaries} frames. The foundation-model comparison is limited to the two open-weight models whose released weights support a linear probe.

\subsection{Broader impact}

The setting TRACE addresses is clinician-supervised ECG diagnosis. The benefit is localizability: a wrong prediction can be attributed to a named pathway, so failure modes can be discussed in clinical language and traced by a developer. The main risks are those common to deployed ECG models: over-reliance by non-expert users, subgroup performance disparities, and misuse under distribution shift, and they apply with particular force to a model whose routing-level auditability might be mistaken for validity. What the architecture guarantees is the wiring; what deployment on it would additionally require is site-specific calibration, subgroup auditing, and abstention policies.


\section{Conclusion}\label{sec:conclusion}

We presented TRACE, an ECG autoencoder whose structure and decision-pathway auditability come from one source: a clinically grounded diagnostic ontology. The ontology is instantiated as a partitioned clinical latent space with hard routing and an orthogonality penalty; the instantiation is then verified by probe, erasure, and perturbation analyses. On PTB-XL and Georgia, TRACE is more accurate than both the unconstrained classifiers and the open-weight foundation models, despite training from scratch on far less data. On the CPSC2018 cohort the same framework transfers with only the routing table re-specified. Every diagnosis is routed through a pathway fixed before training, reconstruction is accurate, and interventions at the mechanism layer move the re-decoded waveform.

The paper's central thesis is procedural. Where reliable domain knowledge exists but the supervision needed to learn it does not, that knowledge can be specified into the model's skeleton and the skeleton then verified explicitly. TRACE demonstrates the thesis for ECG diagnosis, with the ontology fixing the routing before training. The routing-contract protocol of probe, erasure, and perturbation analyses applies to models that claim structured representations. Routing-level transparency follows from construction and representation-level evidence from measurement, and extending both from subspace semantics to feature-level causally grounded explanations is the next step.

\section*{Declaration of competing interest}
The authors declare that they have no known competing financial interests or personal relationships that could have appeared to influence the work reported in this paper.

\section*{Funding}
This work was not supported by any external funding, grants, or sponsored projects.

\section*{Declaration of generative AI and AI-assisted technologies in the manuscript preparation process}
During the preparation of this work the authors used DeepSeek to support language editing and drafting. After using this tool, the authors reviewed and edited the content as needed and take full responsibility for the content of the published article.

\section*{Data availability}
All datasets used in this study are publicly available through PhysioNet: PTB-XL and Georgia (12-lead diagnostic cohorts) and CPSC2018 (nine labels with no structural class). The preprocessing pipeline and evaluation protocol are described in Section~\ref{sec:setup}. The reference implementation is available at \url{https://github.com/R4nzer/TRACE}.

\printcredits

\bibliographystyle{elsarticle-num}
\bibliography{references}

\begin{thebibliography}{10}
\expandafter\ifx\csname url\endcsname\relax
  \def\url#1{\texttt{#1}}\fi
\expandafter\ifx\csname urlprefix\endcsname\relax\def\urlprefix{URL }\fi
\expandafter\ifx\csname href\endcsname\relax
  \def\href#1#2{#2} \def\path#1{#1}\fi

\bibitem{siontis2021artificial}
K.~C. Siontis, P.~A. Noseworthy, Z.~I. Attia, P.~A. Friedman, Artificial
  intelligence-enhanced electrocardiography in cardiovascular disease
  management, Nature Reviews Cardiology 18~(7) (2021) 465--478.
\newblock \href {https://doi.org/10.1038/s41569-020-00503-2}
  {\path{doi:10.1038/s41569-020-00503-2}}.

\bibitem{hannun2019cardiologist}
A.~Y. Hannun, P.~Rajpurkar, M.~Haghpanahi, G.~H. Tison, C.~Bourn, M.~P.
  Turakhia, A.~Y. Ng, Cardiologist-level arrhythmia detection and
  classification in ambulatory electrocardiograms using a deep neural network,
  Nature medicine 25~(1) (2019) 65--69.
\newblock \href {https://doi.org/10.1038/s41591-018-0268-3}
  {\path{doi:10.1038/s41591-018-0268-3}}.

\bibitem{ECGFounder2025}
J.~Li, A.~D. Aguirre, V.~Moura~Junior, J.~Jin, C.~Liu, L.~Zhong, C.~Sun,
  G.~Clifford, M.~B. Westover, S.~Hong, An electrocardiogram foundation model
  built on over 10 million recordings, NEJM AI 2~(7) (2025).
\newblock \href {https://doi.org/10.1056/aioa2401033}
  {\path{doi:10.1056/aioa2401033}}.

\bibitem{HeartLang2025}
J.~Jin, H.~Wang, H.~Li, J.~Li, J.~Pan, S.~Hong, Reading your heart: Learning
  ecg words and sentences via pre-training ecg language model, in:
  International Conference on Learning Representations (ICLR), 2025,
  arXiv:2502.10707.

\bibitem{d2022underspecification}
A.~D'Amour, K.~Heller, D.~Moldovan, B.~Adlam, B.~Alipanahi, A.~Beutel, C.~Chen,
  J.~Deaton, J.~Eisenstein, M.~D. Hoffman, et~al., Underspecification presents
  challenges for credibility in modern machine learning, Journal of Machine
  Learning Research 23~(226) (2022) 1--61.
\newblock \href {https://doi.org/10.48550/arXiv.2011.03395}
  {\path{doi:10.48550/arXiv.2011.03395}}.

\bibitem{geirhos2020shortcut}
R.~Geirhos, J.-H. Jacobsen, C.~Michaelis, R.~Zemel, W.~Brendel, M.~Bethge,
  F.~A. Wichmann, Shortcut learning in deep neural networks, Nature Machine
  Intelligence 2~(11) (2020) 665--673.
\newblock \href {https://doi.org/10.1038/s42256-020-00257-z}
  {\path{doi:10.1038/s42256-020-00257-z}}.

\bibitem{rudin2019stop}
C.~Rudin, Stop explaining black box machine learning models for high stakes
  decisions and use interpretable models instead, Nature machine intelligence
  1~(5) (2019) 206--215.
\newblock \href {https://doi.org/10.1038/s42256-019-0048-x}
  {\path{doi:10.1038/s42256-019-0048-x}}.

\bibitem{ghassemi2021false}
M.~Ghassemi, L.~Oakden-Rayner, A.~L. Beam, The false hope of current approaches
  to explainable artificial intelligence in health care, The lancet digital
  health 3~(11) (2021) e745--e750.
\newblock \href {https://doi.org/10.1016/S2589-7500(21)00208-9}
  {\path{doi:10.1016/S2589-7500(21)00208-9}}.

\bibitem{koh2020concept}
P.~W. Koh, T.~Nguyen, Y.~S. Tang, S.~Mussmann, E.~Pierson, B.~Kim, P.~Liang,
  Concept bottleneck models, in: International Conference on Machine Learning,
  PMLR, 2020, pp. 5338--5348.
\newblock \href {https://doi.org/10.48550/arXiv.2007.04612}
  {\path{doi:10.48550/arXiv.2007.04612}}.

\bibitem{ribeiro2020automatic}
A.~H. Ribeiro, M.~H. Ribeiro, G.~M. Paix{\~a}o, D.~M. Oliveira, P.~R. Gomes,
  J.~A. Canazart, M.~P. Ferreira, C.~R. Andersson, P.~W. Macfarlane,
  W.~Meira~Jr, T.~B. Sch{\"o}n, A.~L.~P. Ribeiro, Automatic diagnosis of the
  12-lead ecg using a deep neural network, Nature communications 11~(1) (2020)
  1760.
\newblock \href {https://doi.org/10.1038/s41467-020-15432-4}
  {\path{doi:10.1038/s41467-020-15432-4}}.

\bibitem{wang2017time}
Z.~Wang, W.~Yan, T.~Oates, Time series classification from scratch with deep
  neural networks: A strong baseline, in: 2017 International Joint Conference
  on Neural Networks (IJCNN), IEEE, 2017, pp. 1578--1585.
\newblock \href {https://doi.org/10.1109/IJCNN.2017.7966039}
  {\path{doi:10.1109/IJCNN.2017.7966039}}.

\bibitem{ismail2020inceptiontime}
H.~Ismail~Fawaz, B.~Lucas, G.~Forestier, C.~Pelletier, D.~F. Schmidt, J.~Weber,
  G.~I. Webb, L.~Idoumghar, P.-A. Muller, F.~Petitjean, Inceptiontime: Finding
  alexnet for time series classification, Data Mining and Knowledge Discovery
  34~(6) (2020) 1936--1962.
\newblock \href {https://doi.org/10.1007/s10618-020-00710-y}
  {\path{doi:10.1007/s10618-020-00710-y}}.

\bibitem{vaswani2017attention}
A.~Vaswani, N.~Shazeer, N.~Parmar, J.~Uszkoreit, L.~Jones, A.~N. Gomez,
  L.~Kaiser, I.~Polosukhin, Attention is all you need, in: Advances in Neural
  Information Processing Systems 30, 2017, pp. 5998--6008.
\newblock \href {https://doi.org/10.48550/arXiv.1706.03762}
  {\path{doi:10.48550/arXiv.1706.03762}}.

\bibitem{li2023dualscale}
Y.~Li, G.~Wang, Z.~Xia, W.~Yang, L.~Sun, A dual-scale lead-separated
  transformer for {ECG} classification, in: 2023 45th Annual International
  Conference of the IEEE Engineering in Medicine \& Biology Society (EMBC),
  2023, pp. 1--4.
\newblock \href {https://doi.org/10.1109/EMBC40787.2023.10340556}
  {\path{doi:10.1109/EMBC40787.2023.10340556}}.

\bibitem{kiyasseh2021clocs}
D.~Kiyasseh, T.~Zhu, D.~A. Clifton, Clocs: Contrastive learning of cardiac
  signals across space, time, and patients, in: International conference on
  machine learning, PMLR, 2021, pp. 5606--5615.
\newblock \href {https://doi.org/10.48550/arXiv.2005.13249}
  {\path{doi:10.48550/arXiv.2005.13249}}.

\bibitem{chen2020simple}
T.~Chen, S.~Kornblith, M.~Norouzi, G.~Hinton, A simple framework for
  contrastive learning of visual representations, in: Proceedings of the 37th
  International Conference on Machine Learning, 2020, pp. 1597--1607.
\newblock \href {https://doi.org/10.48550/arXiv.2002.05709}
  {\path{doi:10.48550/arXiv.2002.05709}}.

\bibitem{mehari2022self}
T.~Mehari, N.~Strodthoff, Self-supervised representation learning from 12-lead
  ecg data, Computers in biology and medicine 141 (2022) 105114.
\newblock \href {https://doi.org/10.1016/j.compbiomed.2021.105114}
  {\path{doi:10.1016/j.compbiomed.2021.105114}}.

\bibitem{he2022masked}
K.~He, X.~Chen, S.~Xie, Y.~Li, P.~Dollar, R.~Girshick, Masked autoencoders are
  scalable vision learners, in: Proceedings of the IEEE/CVF Conference on
  Computer Vision and Pattern Recognition, 2022, pp. 15979--15988.
\newblock \href {https://doi.org/10.1109/CVPR52688.2022.01553}
  {\path{doi:10.1109/CVPR52688.2022.01553}}.

\bibitem{zhang2023maefe}
H.~Zhang, W.~Liu, J.~Shi, S.~Chang, H.~Wang, J.~He, Q.~Huang, {MaeFE}: Masked
  autoencoders family of electrocardiogram for self-supervised pretraining and
  transfer learning, IEEE Transactions on Instrumentation and Measurement 72
  (2023) 1--15.
\newblock \href {https://doi.org/10.1109/TIM.2022.3228267}
  {\path{doi:10.1109/TIM.2022.3228267}}.

\bibitem{pham2025maskedecgtext}
H.~M. Pham, A.~Saeed, D.~Ma, Boosting masked {ECG}-text auto-encoders as
  discriminative learners, in: International Conference on Machine Learning
  (ICML), Vol. 267, 2025, pp. 49277--49291, arXiv:2410.02131.

\bibitem{abbaspourazad2024wearable}
S.~Abbaspourazad, O.~Elachqar, A.~C. Miller, S.~Emrani, U.~Nallasamy,
  I.~Shapiro, Large-scale training of foundation models for wearable
  biosignals, in: International Conference on Learning Representations (ICLR),
  2024, arXiv:2312.05409.

\bibitem{mckeen2025ecgfm}
K.~McKeen, S.~Masood, A.~Toma, B.~Rubin, B.~Wang, {ECG-FM}: an open
  electrocardiogram foundation model, JAMIA Open 8~(5) (2025) ooaf122.
\newblock \href {https://doi.org/10.1093/jamiaopen/ooaf122}
  {\path{doi:10.1093/jamiaopen/ooaf122}}.

\bibitem{vaid2023foundational}
A.~Vaid, J.~Jiang, A.~Sawant, S.~Lerakis, E.~Argulian, Y.~Ahuja, J.~Lampert,
  A.~Charney, H.~Greenspan, J.~Narula, B.~Glicksberg, G.~N. Nadkarni, A
  foundational vision transformer improves diagnostic performance for
  electrocardiograms, npj Digital Medicine 6~(1) (2023) 108.
\newblock \href {https://doi.org/10.1038/s41746-023-00840-9}
  {\path{doi:10.1038/s41746-023-00840-9}}.

\bibitem{papastathopoulos2026physics}
A.~Papastathopoulos-Katsaros, A.~Stavrianidi, Z.~Liu, Physics-informed deep
  learning for false ventricular tachycardia alarm reduction in the {ICU}, in:
  Computing in Cardiology (CinC), 2026, arXiv:2609.08992.

\bibitem{na2024guiding}
Y.~Na, M.~Park, Y.~Tae, S.~Joo, Guiding masked representation learning to
  capture spatio-temporal relationship of electrocardiogram, in: International
  Conference on Learning Representations (ICLR), 2024, arXiv:2402.09450.

\bibitem{margeloiu2021concept}
A.~Margeloiu, M.~Ashman, U.~Bhatt, Y.~Chen, M.~Jamnik, A.~Weller, Do concept
  bottleneck models learn as intended?, arXiv preprint arXiv:2105.04289ICLR
  2021 Workshop on Responsible AI (2021).

\bibitem{oikarinen2023labelfree}
T.~Oikarinen, S.~Das, L.~M. Nguyen, T.-W. Weng, Label-free concept bottleneck
  models, in: International Conference on Learning Representations (ICLR),
  2023, arXiv:2304.06129.

\bibitem{espinosazarlenga2022concept}
M.~Espinosa~Zarlenga, P.~Barbiero, G.~Ciravegna, G.~Marra, F.~Giannini,
  M.~Diligenti, Z.~Shams, F.~Precioso, S.~Melacci, A.~Weller, P.~Li{\'o},
  M.~Jamnik, Concept embedding models: Beyond the accuracy-explainability
  trade-off, in: Advances in Neural Information Processing Systems (NeurIPS),
  Vol.~35, 2022, pp. 21400--21413, arXiv:2209.09056.

\bibitem{xu2024energy}
X.~Xu, Y.~Qin, L.~Mi, H.~Wang, X.~Li, Energy-based concept bottleneck models:
  Unifying prediction, concept intervention, and probabilistic interpretations,
  in: International Conference on Learning Representations (ICLR), 2024,
  arXiv:2401.14142.

\bibitem{nauta2023pipnet}
M.~Nauta, J.~Schl{\"o}tterer, M.~van Keulen, C.~Seifert, {PIP-Net}: Patch-based
  intuitive prototypes for interpretable image classification, in: Proceedings
  of the IEEE/CVF Conference on Computer Vision and Pattern Recognition (CVPR),
  2023, pp. 2744--2753.
\newblock \href {https://doi.org/10.1109/CVPR52729.2023.00269}
  {\path{doi:10.1109/CVPR52729.2023.00269}}.

\bibitem{sethi2025protoecgnet}
S.~Sethi, D.~Chen, T.~Statchen, M.~C. Burkhart, N.~Bhandari, B.~Ramadan,
  B.~Beaulieu-Jones, {ProtoECGNet}: Case-based interpretable deep learning for
  multi-label {ECG} classification with contrastive learning, in: Proceedings
  of the 10th Machine Learning for Healthcare Conference (MLHC), Vol. 298 of
  Proceedings of Machine Learning Research, 2025, arXiv:2504.08713.

\bibitem{elazar2021amnesic}
Y.~Elazar, S.~Ravfogel, A.~Jacovi, Y.~Goldberg, Amnesic probing: Behavioral
  explanation with amnesic counterfactuals, Transactions of the Association for
  Computational Linguistics 9 (2021) 160--175.
\newblock \href {https://doi.org/10.1162/tacl_a_00359}
  {\path{doi:10.1162/tacl_a_00359}}.

\bibitem{belinkov2022probing}
Y.~Belinkov, Probing classifiers: Promises, shortcomings, and advances,
  Computational Linguistics 48~(1) (2022) 207--219.
\newblock \href {https://doi.org/10.1162/coli_a_00422}
  {\path{doi:10.1162/coli_a_00422}}.

\bibitem{yisimitila2025interpretable}
T.~Yisimitila, C.~Wang, M.~Hou, M.~Maimaitiniyazi, Z.~Aili, T.~Liu, H.~Tan,
  N.~Chellamani, M.~Shanmuganathan, M.~Liu, M.~Nijiati, Bridging clinical
  knowledge and ai: an interpretable transformer framework for ecg diagnosis,
  npj Digital Medicine 9~(1) (2026) 41.
\newblock \href {https://doi.org/10.1038/s41746-025-02215-8}
  {\path{doi:10.1038/s41746-025-02215-8}}.

\bibitem{strodthoff2021deep}
N.~Strodthoff, P.~Wagner, T.~Schaeffter, W.~Samek, Deep learning for ecg
  analysis: Benchmarks and insights from ptb-xl, IEEE Journal of Biomedical and
  Health Informatics 25~(5) (2021) 1519--1528.
\newblock \href {https://doi.org/10.1109/JBHI.2020.3022989}
  {\path{doi:10.1109/JBHI.2020.3022989}}.

\bibitem{taleban2026explainable}
A.~Taleban, R.~Sparapani, P.~Noffke, S.~Zlochiver, Q.~Lu, M.~E. Widlansky,
  J.~Luo, Explainable artificial intelligence in electrocardiography: A
  systematic review, Biomedical Signal Processing and Control 114 (2026)
  109325.
\newblock \href {https://doi.org/10.1016/j.bspc.2025.109325}
  {\path{doi:10.1016/j.bspc.2025.109325}}.

\bibitem{subbaswamy2020development}
A.~Subbaswamy, S.~Saria, From development to deployment: dataset shift,
  causality, and shift-stable models in health ai, Biostatistics 21~(2) (2020)
  345--352.
\newblock \href {https://doi.org/10.1093/biostatistics/kxz041}
  {\path{doi:10.1093/biostatistics/kxz041}}.

\bibitem{karpatne2024knowledgeguided}
A.~Karpatne, X.~Jia, V.~Kumar, Knowledge-guided machine learning: Current
  trends and future prospects, arXiv preprint arXiv:2403.15989 (2024).

\bibitem{badreddine2022logic}
S.~Badreddine, A.~d'Avila Garcez, L.~Serafini, M.~Spranger, Logic tensor
  networks, Artificial Intelligence 303 (2022) 103649.
\newblock \href {https://doi.org/10.1016/j.artint.2021.103649}
  {\path{doi:10.1016/j.artint.2021.103649}}.

\bibitem{raissi2019physics}
M.~Raissi, P.~Perdikaris, G.~E. Karniadakis, Physics-informed neural networks:
  A deep learning framework for solving forward and inverse problems involving
  nonlinear partial differential equations, Journal of Computational Physics
  378 (2019) 686--707.
\newblock \href {https://doi.org/10.1016/j.jcp.2018.10.045}
  {\path{doi:10.1016/j.jcp.2018.10.045}}.

\bibitem{higgins2017betavae}
I.~Higgins, L.~Matthey, A.~Pal, C.~Burgess, X.~Glorot, M.~Botvinick,
  S.~Mohamed, A.~Lerchner, beta-{VAE}: Learning basic visual concepts with a
  constrained variational framework, in: International Conference on Learning
  Representations (ICLR), 2017, openReview: Sy2fzU9gl.

\bibitem{kim2018disentangling}
H.~Kim, A.~Mnih, Disentangling by factorising, in: International Conference on
  Machine Learning (ICML), 2018, pp. 2649--2658, arXiv:1802.05983.

\bibitem{chen2018isolating}
R.~T.~Q. Chen, X.~Li, R.~B. Grosse, D.~Duvenaud, Isolating sources of
  disentanglement in variational autoencoders, in: Advances in Neural
  Information Processing Systems (NeurIPS), 2018, pp. 2610--2620.
\newblock \href {https://doi.org/10.48550/arXiv.1802.04942}
  {\path{doi:10.48550/arXiv.1802.04942}}.

\bibitem{trauble2021disentangled}
F.~Tr{\"a}uble, E.~Creager, N.~Kilbertus, F.~Locatello, A.~Dittadi, A.~Goyal,
  B.~Sch{\"o}lkopf, S.~Bauer, On disentangled representations learned from
  correlated data, in: International Conference on Machine Learning, PMLR,
  2021, pp. 10401--10412.
\newblock \href {https://doi.org/10.48550/arXiv.2006.07886}
  {\path{doi:10.48550/arXiv.2006.07886}}.

\bibitem{scholkopf2021toward}
B.~Sch{\"o}lkopf, F.~Locatello, S.~Bauer, N.~R. Ke, N.~Kalchbrenner, A.~Goyal,
  Y.~Bengio, Toward causal representation learning, Proceedings of the IEEE
  109~(5) (2021) 612--634.
\newblock \href {https://doi.org/10.1109/JPROC.2021.3058954}
  {\path{doi:10.1109/JPROC.2021.3058954}}.

\bibitem{locatello2019challenging}
F.~Locatello, S.~Bauer, M.~Lucic, G.~R{\"a}tsch, S.~Gelly, B.~Sch{\"o}lkopf,
  O.~Bachem, Challenging common assumptions in the unsupervised learning of
  disentangled representations, in: International Conference on Machine
  Learning (ICML), Vol.~97, 2019, pp. 4114--4124, arXiv:1811.12359.

\bibitem{khemakhem2020variational}
I.~Khemakhem, D.~P. Kingma, R.~P. Monti, A.~Hyv{\"a}rinen, Variational
  autoencoders and nonlinear {ICA}: A unifying framework, in: International
  Conference on Artificial Intelligence and Statistics (AISTATS), 2020, pp.
  2207--2217, arXiv:1907.04809.

\bibitem{matabuena2019heartrate}
M.~Matabuena, J.~C. Vidal, P.~R. Hayes, M.~Saavedra-Garcia, F.~Huelin~Trillo,
  Application of functional data analysis for the prediction of maximum heart
  rate, IEEE Access 7 (2019) 121841--121852.
\newblock \href {https://doi.org/10.1109/ACCESS.2019.2938466}
  {\path{doi:10.1109/ACCESS.2019.2938466}}.

\bibitem{matabuena2024multilevel}
M.~Matabuena, C.~M. Crainiceanu, Multilevel functional distributional models
  with applications to continuous glucose monitoring in diabetes clinical
  trials, The Annals of Applied Statistics 20~(1) (2026).
\newblock \href {https://doi.org/10.1214/26-aoas2139}
  {\path{doi:10.1214/26-aoas2139}}.

\bibitem{ghosal2025functional}
R.~Ghosal, M.~Matabuena, S.~K. Ghosh, Functional time transformation model with
  applications to digital health, Computational Statistics \& Data Analysis 207
  (2025) 108131.
\newblock \href {https://doi.org/10.1016/j.csda.2025.108131}
  {\path{doi:10.1016/j.csda.2025.108131}}.

\bibitem{marriott2008practical}
G.~S. Wagner, D.~G. Strauss, Marriott's Practical Electrocardiography, 12th
  Edition, Lippincott Williams \& Wilkins, 2014.

\bibitem{guyton2020textbook}
J.~E. Hall, M.~E. Hall, Guyton and Hall Textbook of Medical Physiology, 14th
  Edition, Elsevier, 2020.

\bibitem{wellens200640}
H.~J.~J. Wellens, A.~P.~M. Gorgels, P.~A. F.~M. Doevendans, The ECG in Acute
  Myocardial Infarction and Unstable Angina: Diagnosis and Risk Stratification,
  Developments in Cardiovascular Medicine, Springer, 2002.
\newblock \href {https://doi.org/10.1007/b101885} {\path{doi:10.1007/b101885}}.

\bibitem{braunwald2018heart}
D.~P. Zipes, P.~Libby, R.~O. Bonow, D.~L. Mann, G.~F. Tomaselli, Braunwald's
  Heart Disease: A Textbook of Cardiovascular Medicine, 11th Edition, Elsevier,
  2019.

\bibitem{josephson2015clinical}
M.~E. Josephson, Josephson's Clinical Cardiac Electrophysiology: Techniques and
  Interpretations, 5th Edition, Wolters Kluwer Health, 2016.

\bibitem{thygesen2012universal}
K.~Thygesen, J.~S. Alpert, A.~S. Jaffe, M.~L. Simoons, B.~R. Chaitman, H.~D.
  White, H.~A. Katus, F.~S. Apple, B.~Lindahl, D.~A. Morrow, P.~M. Clemmensen,
  et~al., Third universal definition of myocardial infarction, Journal of the
  American College of Cardiology 60~(16) (2012) 1581--1598.
\newblock \href {https://doi.org/10.1016/j.jacc.2012.08.001}
  {\path{doi:10.1016/j.jacc.2012.08.001}}.

\bibitem{wagner2020ptbxl}
P.~Wagner, N.~Strodthoff, R.-D. Bousseljot, D.~Kreiseler, F.~I. Lunze,
  W.~Samek, T.~Schaeffter,
  \href{https://doi.org/10.1038/s41597-020-0495-6}{{PTB-XL, a large publicly
  available electrocardiography dataset}}, Scientific Data 7~(1) (2020) 154.
\newblock \href {https://doi.org/10.1038/s41597-020-0495-6}
  {\path{doi:10.1038/s41597-020-0495-6}}.
\newline\urlprefix\url{https://doi.org/10.1038/s41597-020-0495-6}

\bibitem{perezalday2020challenge}
E.~A. Perez~Alday, A.~Gu, A.~J. Shah, C.~Robichaux, A.-K.~I. Wong, C.~Liu,
  F.~Liu, A.~Bahrami~Rad, A.~Elola, S.~Seyedi, Q.~Li, A.~Sharma, G.~D.
  Clifford, M.~A. Reyna,
  \href{http://doi.org/10.1088/1361-6579/abc960}{{Classification of 12-lead
  ECGs: The PhysioNet/Computing in Cardiology Challenge 2020}}, Physiological
  Measurement 41~(12) (2020) 124003.
\newblock \href {https://doi.org/10.1088/1361-6579/abc960}
  {\path{doi:10.1088/1361-6579/abc960}}.
\newline\urlprefix\url{http://doi.org/10.1088/1361-6579/abc960}

\bibitem{perezalday2022challengephysionet}
E.~A. Perez~Alday, A.~Gu, A.~Shah, C.~Liu, A.~Sharma, S.~Seyedi,
  A.~Bahrami~Rad, M.~Reyna, G.~Clifford,
  \href{https://doi.org/10.13026/dvyd-kd57}{{Classification of 12-lead ECGs:
  The PhysioNet/Computing in Cardiology Challenge 2020 (version 1.0.2)}},
  PhysioNet, {RRID}:SCR\_007345 (2022).
\newblock \href {https://doi.org/10.13026/dvyd-kd57}
  {\path{doi:10.13026/dvyd-kd57}}.
\newline\urlprefix\url{https://doi.org/10.13026/dvyd-kd57}

\bibitem{liu2018open}
F.~Liu, C.~Liu, L.~Zhao, X.~Zhang, X.~Wu, X.~Xu, Y.~Liu, C.~Ma, S.~Wei, Z.~He,
  J.~Li, E.~Ng~Yin~Kwee, An open access database for evaluating the algorithms
  of electrocardiogram rhythm and morphology abnormality detection, Journal of
  Medical Imaging and Health Informatics 8~(7) (2018) 1368--1373.
\newblock \href {https://doi.org/10.1166/jmihi.2018.2442}
  {\path{doi:10.1166/jmihi.2018.2442}}.

\bibitem{kingma2014autoencoding}
D.~P. Kingma, M.~Welling, Auto-encoding variational bayes, in: International
  Conference on Learning Representations (ICLR), 2014, arXiv:1312.6114.

\bibitem{goldberger2000physiobank}
A.~L. Goldberger, L.~A.~N. Amaral, L.~Glass, J.~M. Hausdorff, P.~C. Ivanov,
  R.~G. Mark, J.~E. Mietus, G.~B. Moody, C.-K. Peng, H.~E. Stanley,
  {PhysioBank, PhysioToolkit, and PhysioNet: Components of a New Research
  Resource for Complex Physiologic Signals}, Circulation 101~(23) (2000)
  e215--e220, {RRID}:SCR\_007345.
\newblock \href {https://doi.org/10.1161/01.CIR.101.23.e215}
  {\path{doi:10.1161/01.CIR.101.23.e215}}.

\bibitem{wagner2020ptbxlphysionet}
P.~Wagner, N.~Strodthoff, R.-D. Bousseljot, W.~Samek, T.~Schaeffter,
  \href{https://doi.org/10.13026/x4td-x982}{{PTB-XL, a large publicly available
  electrocardiography dataset (version 1.0.1)}}, PhysioNet, {RRID}:SCR\_007345
  (2020).
\newblock \href {https://doi.org/10.13026/x4td-x982}
  {\path{doi:10.13026/x4td-x982}}.
\newline\urlprefix\url{https://doi.org/10.13026/x4td-x982}

\bibitem{collins2024tripodai}
G.~S. Collins, K.~G.~M. Moons, P.~Dhiman, R.~D. Riley, A.~L. Beam,
  B.~Van~Calster, M.~Ghassemi, X.~Liu, J.~B. Reitsma, M.~van Smeden, et~al.,
  {TRIPOD+AI} statement: updated guidance for reporting clinical prediction
  models that use regression or machine learning methods, BMJ 385 (2024)
  e078378.
\newblock \href {https://doi.org/10.1136/bmj-2023-078378}
  {\path{doi:10.1136/bmj-2023-078378}}.

\bibitem{varoquaux2022machine}
G.~Varoquaux, V.~Cheplygina, Machine learning for medical imaging:
  methodological failures and recommendations for the future, npj Digital
  Medicine 5~(1) (2022) 48.
\newblock \href {https://doi.org/10.1038/s41746-022-00592-y}
  {\path{doi:10.1038/s41746-022-00592-y}}.

\bibitem{bogatinovski2022comprehensive}
J.~Bogatinovski, L.~Todorovski, S.~D{\v{z}}eroski, D.~Kocev, Comprehensive
  comparative study of multi-label classification methods, Expert Systems with
  Applications 203 (2022) 117215.
\newblock \href {https://doi.org/10.1016/j.eswa.2022.117215}
  {\path{doi:10.1016/j.eswa.2022.117215}}.

\bibitem{brehmer2022weakly}
J.~Brehmer, P.~de~Haan, P.~Lippe, T.~Cohen, Weakly supervised causal
  representation learning, in: Advances in Neural Information Processing
  Systems (NeurIPS), 2022, pp. 38319--38331, arXiv:2203.16437.

\bibitem{feuerriegel2024causal}
S.~Feuerriegel, D.~Frauen, V.~Melnychuk, J.~Schweisthal, K.~Hess, A.~Curth,
  S.~Bauer, N.~Kilbertus, I.~S. Kohane, M.~van~der Schaar, Causal machine
  learning for predicting treatment outcomes, Nature Medicine 30~(4) (2024)
  958--968.
\newblock \href {https://doi.org/10.1038/s41591-024-02902-1}
  {\path{doi:10.1038/s41591-024-02902-1}}.

\bibitem{vancalster2019calibration}
B.~Van~Calster, D.~J. McLernon, M.~van Smeden, L.~Wynants, E.~W. Steyerberg,
  Calibration: the achilles heel of predictive analytics, BMC Medicine 17~(1)
  (2019) 230.
\newblock \href {https://doi.org/10.1186/s12916-019-1466-7}
  {\path{doi:10.1186/s12916-019-1466-7}}.

\bibitem{guo2017calibration}
C.~Guo, G.~Pleiss, Y.~Sun, K.~Q. Weinberger, On calibration of modern neural
  networks, in: International Conference on Machine Learning (ICML), 2017, pp.
  1321--1330, arXiv:1706.04599.

\bibitem{portela2025conformal}
A.~Portela, J.~R. Banga, M.~Matabuena, Conformal prediction for uncertainty
  quantification in dynamic biological systems, PLOS Computational Biology
  21~(5) (2025) e1013098.
\newblock \href {https://doi.org/10.1371/journal.pcbi.1013098}
  {\path{doi:10.1371/journal.pcbi.1013098}}.

\bibitem{lugosi2024uncertainty}
G.~Lugosi, M.~Matabuena, Uncertainty quantification in metric spaces, arXiv
  preprint arXiv:2405.05110 (2024).

\bibitem{xu2023conformal}
C.~Xu, Y.~Xie, Conformal prediction for time series, IEEE Transactions on
  Pattern Analysis and Machine Intelligence 45~(10) (2023) 11575--11587.
\newblock \href {https://doi.org/10.1109/TPAMI.2023.3272339}
  {\path{doi:10.1109/TPAMI.2023.3272339}}.

\bibitem{yu2025trustworthy}
B.~Yu, Y.~Liu, X.~Wu, J.~Ren, Z.~Zhao, Trustworthy diagnosis of
  electrocardiography signals based on out-of-distribution detection, PLOS ONE
  20~(2) (2025) e0317900.
\newblock \href {https://doi.org/10.1371/journal.pone.0317900}
  {\path{doi:10.1371/journal.pone.0317900}}.

\end{thebibliography}

\end{document}